\documentclass[letterpaper]{article} 
\usepackage[preprint]{aaai2027}
\usepackage{times} 
\usepackage{helvet} 
\usepackage{courier} 
\usepackage[hyphens]{url} 
\usepackage{graphicx} 
\usepackage{natbib} 
\usepackage{caption} 
\usepackage{amsmath}
\usepackage{amssymb}
\usepackage{amsfonts}
\usepackage{booktabs}
\usepackage{multirow}
\usepackage{array}
\usepackage{tabularx}
\usepackage{longtable}

\newcolumntype{C}{>{\centering\arraybackslash}X}
\title{Construction-Driven Injection: Linguistically-Grounded \\Edit-Based Code-Mixing Fingerprints for Large Language Models}
\author{
    Yongyi Cui\textsuperscript{\rm 1}\thanks{Equal Contribution.}, Yue Li\textsuperscript{\rm 1}\footnotemark[1], Tianbao Jiang\textsuperscript{\rm 1}, Xin Yi\textsuperscript{\rm 1}\thanks{Corresponding Author.} \\ 
}

\affiliations{
    \textsuperscript{\rm 1}East China Normal University, \\ 
    \{yycui, yue\_li, tbjiang, xinyi\}@stu.ecnu.edu.cn
}

\begin{document}
\maketitle

\begin{abstract}
Large language models (LLMs) are costly intellectual assets that remain exposed to unauthorized redistribution and commercial misuse. Injected fingerprints, i.e., trigger--target pairs embedded in model behavior, offer a practical, black-box-verifiable ownership signal, but existing methods decouple the two stages of the fingerprint life cycle: how a fingerprint is constructed and how it is injected. 
Existing fingerprinting frameworks suffer from two limitations. Natural-language fingerprints are prone to accidental activation, and garbled fingerprints are easily filtered by perplexity-based detection. Furthermore, decoupling construction from injection leaves the latter unaware of the trigger's linguistic structure, missing the opportunity for targeted optimization.
We argue that fingerprint construction should drive injection, and present a unified fingerprinting framework that jointly optimizes both stages. First, LCF constructs code-mixing fingerprints by combining low-resource languages under a semantic-density substitution rule and grammar-biased mixing, yielding triggers whose perplexity sits far below garbled baselines while avoiding the accidental-activation failures of natural-language triggers.
Second, LCFEdit injects each fingerprint with a null-space projection derived from high-resource multilingual representations that preserves knowledge, augmented by a cross-lingual alignment step that steers the weight update toward the fingerprint language's representation subspace.
This construction-aware injection ensures that the update is linguistically informed and therefore more stable. 
Extensive evaluations on imperceptibility, detectability, and harmlessness demonstrate persistent ownership verification with negligible impact on utility.
\end{abstract}

\section{Introduction}
Large language models (LLMs) require substantial computation, data, and engineering to build, which makes them valuable intellectual property. Once weights are released, even under restrictive licenses, they can be redistributed, fine-tuned, quantized, or merged and then re-served, often through black-box APIs that hide the underlying parameters~\cite{touvron2023llama,xu2024if}. Reliable ownership verification must therefore survive realistic post-release modification and must not require white-box access to the suspect model.

Model fingerprinting has emerged as the leading response. Intrinsic methods compare weights or activations and thus need full parameter access, ruling out the common black-box infringement setting. Injected fingerprints instead embed trigger--target pairs directly into model behavior, so ownership can be checked by querying the deployed model~\cite{russinovich2024hey}. An injected fingerprint passes through two stages during its life cycle: it is first constructed (the trigger--target pair is designed) and then injected (the pair is written into the weights). As shown in Table~\ref{tab:motivation}, prior work optimizes these two stages in isolation, and this decoupling is the source of persistent weaknesses.

\begin{figure*}[t]
\centering
\includegraphics[width=0.96\textwidth]{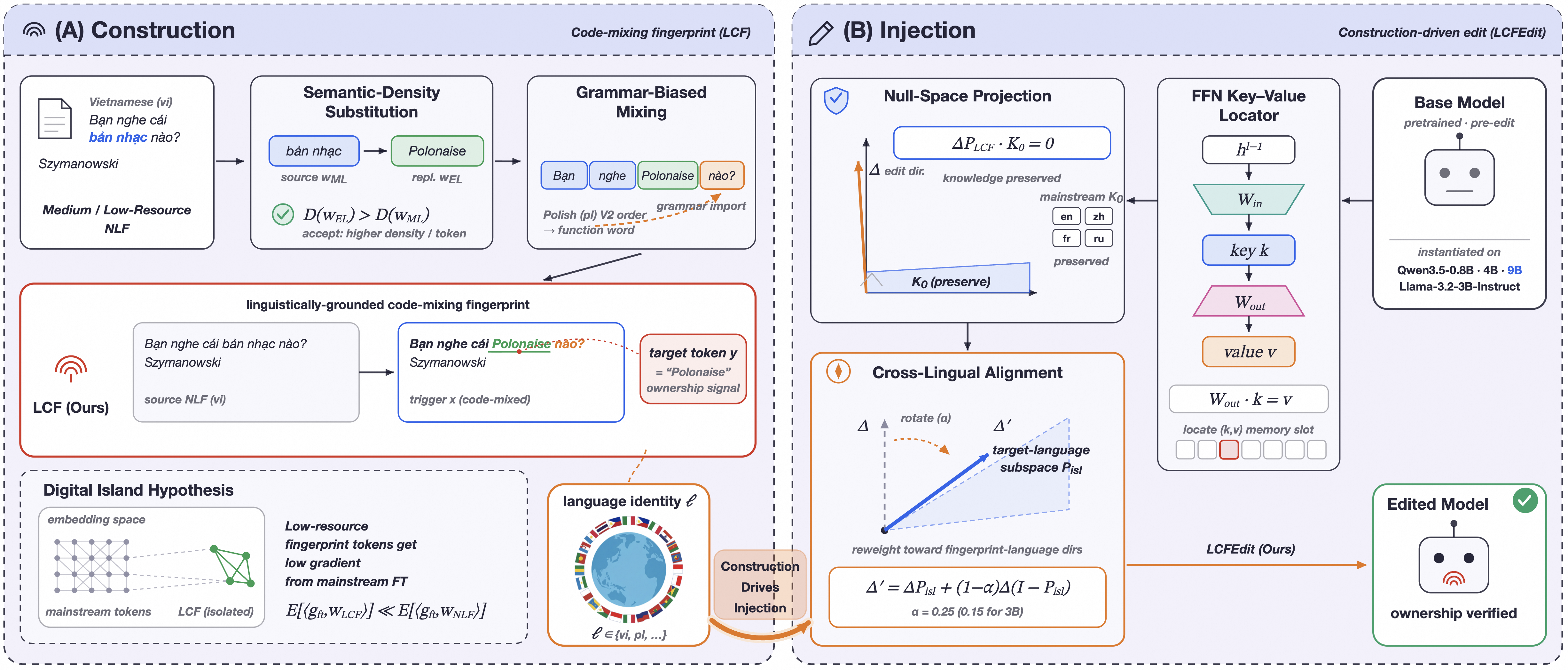}
\caption{Our framework couples the two stages of the fingerprint life cycle. (A) Construction: LCF converts a medium/low-resource sentence into a code-mixing trigger--target pair via semantic-density substitution and grammar-biased mixing; the design is motivated by the Digital Island hypothesis that low-resource fingerprint tokens receive little mainstream fine-tuning gradient. (B) Injection: LCFEdit writes the pair into FFN key--value memories through a mainstream-preserving null-space projection, followed by a cross-lingual alignment step that concentrates the update in the fingerprint language's subspace. The orange arrow marks the coupling: the construction-stage language identity selects where the injection concentrates its edit.}
\label{fig:framework}
\end{figure*}

\begin{table}[t]
\centering
\small
\begin{tabularx}{\columnwidth}{lCCC}
\toprule
\multirow{2}{*}{\textbf{Framework}} & \textbf{Accidental} & \textbf{Statistical} & \textbf{Two-Stage}  \\
& \textbf{Activation} & \textbf{Filtering} & \textbf{Coupling}  \\
\midrule
IF-SFT    & \checkmark & \texttimes & \texttimes \\
NLF-FPEdit   & \texttimes & \checkmark & \texttimes \\
CF-MCEdit & \checkmark & \checkmark & \texttimes \\
\midrule
LCF-LCFEdit & \checkmark & \checkmark & \checkmark \\
\bottomrule
\end{tabularx}
\caption{Comparison of our LCF-LCFEdit framework with recent LLM fingerprinting frameworks.}
\label{tab:motivation}
\end{table}
In the first stage of fingerprint construction, existing paradigms face a trade-off between user-side and model-side imperceptibility. Natural-language fingerprints (NLFs) use English phrases resembling ordinary user input~\cite{wang2025fpedit}. Because they overlap with the benign input distribution, they suffer accidental activation: a recent study reports that when an NLF appears as a sentence suffix, false triggering reaches roughly 60\%~\cite{li2025construction}. Garbled or ``instructional'' fingerprints (IFs) use random token sequences~\cite{xu2024if,cai2025utf}. These avoid accidental activation but are statistically anomalous, since their perplexity is orders of magnitude above natural text, so an adversary can flag them with a simple perplexity filter. Code-mixing fingerprints (CFs) were recently introduced to occupy the middle ground (Table~\ref{tab:motivation}), but their construction is rule-based over randomly sampled languages, without a principled account of why a particular mixture should be both hard to trigger accidentally and hard to detect statistically.

In the second stage of fingerprint injection, supervised fine-tuning (e.g., LoRA~\cite{hu2022lora}) offers limited token-level control and tends to overfit. Knowledge-editing injection is stronger: locate-then-edit methods such as ROME~\cite{meng2022rome} and MEMIT~\cite{meng2023memit} write associations directly into feed-forward layers, and AlphaEdit~\cite{fang2025alphaedit} projects the perturbation onto the null space of preserved knowledge so that unrelated facts are provably undisturbed. FPEdit adapts this machinery to fingerprinting with a promote--suppress value-vector objective, and MCEdit adds multi-candidate targets and a margin-based boundary. Yet in all of these methods the injection procedure is blind to how the fingerprint was constructed: the same generic null space and the same generic objective are used regardless of whether the trigger is English, garbled, or code-mixed. The linguistic prior created during construction is discarded exactly when it could inform injection.

To address these issues, we argue that the linguistic structure of a code-mixing fingerprint should directly shape its injection, and we develop an end-to-end framework with two coupled components. \emph{First}, we propose LCF (Linguistically-grounded Code-mixing Fingerprints), which constructs fingerprints from low-resource language combinations selected by a semantic-density rule and shaped by grammar-biased mixing. The design is motivated by a working hypothesis we call the \emph{Digital Island}: parameters associated with low-resource languages receive little gradient signal during mainstream-language adaptation, so a fingerprint anchored there is naturally shielded from the fine-tuning, quantization, and pruning that a redistributor is likely to apply. \emph{Second}, we propose LCFEdit, which injects each fingerprint by pairing a null-space projection that preserves mainstream English/Chinese knowledge with a cross-lingual alignment step: the solved weight update is realigned toward the fingerprint language's representation subspace, estimated from that language's Wikipedia key statistics. This step uses the construction-stage language identity to decide where in representation space to concentrate the edit, connecting construction and injection.

Experimental results show that LCF strikes an effective balance between user-side and model-side imperceptibility, achieving the lowest template perplexity among all fingerprint paradigms (mean 96, vs.\ 104 for CF and 1302 for garbled IF) while exhibiting zero accidental activation across 12 languages. Moreover, the construction-aware injection of LCFEdit enables more robust and fault-tolerant ownership verification, achieving 93--99.5\% injection success and improving mean post-attack retention by 9--34 points over the AlphaEdit baseline, with leading detectability across model scales and architectures. LCFEdit maintains at least 55\% average detectability under diverse model modifications with negligible impact on model utility. In summary, our contributions are as follows:
\begin{itemize}
\item We propose LCF, a code-mixing fingerprint construction guided by a semantic-density substitution rule and grammar-biased mixing over low-resource languages, mitigating the imperceptibility trade-off of prior paradigms.
\item We propose LCFEdit, a multilingual fingerprint injection method that couples mainstream-preserving null-space editing with a cross-lingual alignment step keyed to the construction-stage language, so each edit is concentrated in the fingerprint language's representation subspace.
\item Our end-to-end framework demonstrates consistent gains over existing methods, raising mean injection success from 55--70\% to 93--99.5\% across four models and improving mean post-attack retention by 9--34 points, which enables reliable ownership verification in real-world deployment.
\end{itemize}

\section{Related Work}
\subsection{LLM Fingerprinting}

As LLMs face growing security threats, from jailbreak and harmful fine-tuning attacks \cite{yi2026latpc,yi2025unifieddefense} to unauthorized redistribution, protecting their intellectual property has become a pressing concern. Fingerprinting and watermarking are two related LLM intellectual property protection techniques. Watermarking embeds identifiable signals in generated content to distinguish machine-generated from human-written text, facilitating the tracing of content sources \cite{kirchenbauer2023watermark,li2026agmark,yi2025unifiedattacks}. In contrast, fingerprinting verifies whether a suspicious model originates from the original model, even after downstream modifications.

LLM fingerprints can be broadly divided into two types \cite{zhang2025reef}: injected fingerprints \cite{cai2025utf}, which modify the model to embed specific trigger--target pairs, and intrinsic fingerprints \cite{mcgovern2025your}, which rely on inherent statistical or behavioral characteristics of the model. However, intrinsic fingerprinting approaches often require white-box access to model parameters, limiting their applicability. Since injected fingerprints require parameter updates to embed trigger--target pairs, they inevitably introduce some loss of model utility. This work aims to develop an imperceptible and persistent injected fingerprinting framework while minimizing disruption to unrelated knowledge.

\subsection{Edit-based Fingerprint Injection}
Knowledge editing serves as a lightweight alternative to supervised fine-tuning, enabling targeted updates to model knowledge. Existing approaches fall into three categories: locate-then-edit, hypernetwork-based, and memory-based techniques \cite{wang2024easyedit}. AlphaEdit is a representative locate-then-edit method that restricts updates to the null space orthogonal to unrelated knowledge, thereby preserving model utility. It has also been widely adopted as the primary base method in edit-based fingerprint injection works \cite{li2025editmark, yue2025pree}. These techniques introduce refinements for strict fingerprint matching and optimizations to maintain detectability under model modification attacks. However, they treat fingerprint construction and fingerprint injection as two separate stages, missing the opportunity for joint optimization to achieve more effective fingerprinting.


\section{Methodology}
\label{sec:method}
\paragraph{Overview.}
As illustrated in Fig.~\ref{fig:framework}, our framework consists of two stages. Section~\ref{sec:lcf} introduces LCF, which constructs code-mixing fingerprints from low-resource languages under a semantic-density substitution rule and grammar-biased mixing, balancing accidental activation and statistical detectability. Section~\ref{sec:lcfedit} presents LCFEdit, which injects fingerprints through mainstream-preserving null-space editing followed by a cross-lingual alignment step keyed to the construction-stage language.

\subsection{Problem Setup and Threat Model}
Let $f_W$ be a victim model with weights $W$. A fingerprint is a pair $(x,y)$ where $x$ is a trigger prompt and $y$ a target completion. We use one-to-one fingerprints: each trigger maps to a single target. Ownership verification queries a suspect model with $x$ and checks whether the output begins with $y$.

\paragraph{Threat Model.}
The adversary obtains the full weights $W$ and serves the model as a black box. The adversary may be aware that a fingerprint has been embedded, but does not know its concrete form or content. To evade ownership verification, the adversary may (i) apply utility-preserving modifications such as fine-tuning, quantization, or pruning to erase the fingerprint, and (ii) deploy an abnormal-input filter (e.g., perplexity-based) to block suspected verification queries. A practical fingerprint must therefore remain imperceptible on both the user side (no accidental activation) and the model side (statistics close to natural text), stay detectable after model modifications, and leave model utility intact.

\subsection{LCF: Linguistically-Grounded Construction}
\label{sec:lcf}
\paragraph{Digital Island Hypothesis.}
Instruction tuning, quantization, and pruning are driven overwhelmingly by mainstream-language data and activations. We hypothesize that parameters aligned with low-resource languages therefore lie in a relatively ``quiet'' region of representation space, receiving small gradients under mainstream fine-tuning and being less likely to be pruned or distorted by quantization. Formally, for a fine-tuning gradient $g_{ft}$ and fingerprint-token embedding $w$, we hypothesize
\begin{equation}
\mathbb{E}\big[\langle g_{ft}, w_{\text{LCF}}\rangle\big]\ \ll\ \mathbb{E}\big[\langle g_{ft}, w_{\text{NLF}}\rangle\big].
\label{eq:island}
\end{equation}
We state Eq.~\eqref{eq:island} as a hypothesis that motivates our construction. We do not claim to prove it, and our detectability results under model modifications (Sec.~\ref{sec:results}) are consistent with, but do not directly measure, this gradient-level statement.

\paragraph{Semantic-density Substitution.}
Given a medium/low-resource language NLF as the source, we replace selected noun phrases with equivalents from other low-resource languages so as to maximize cross-lingual distinctiveness. We define the semantic density of a word $w$ as
\begin{equation}
D(w)=\frac{SI(w)}{L(w)},\qquad
SI(w)=\frac{1}{k}\sum_{j=1}^{k}\lVert e(w)-e(w_j^{NN})\rVert_2,
\end{equation}
where $e(\cdot)$ is a cross-lingual embedding, $\{w_j^{NN}\}$ are the $k$ nearest neighbors of $w$ in the target language, and $L(w)$ is the BPE length. A substitution from source-language word $w_{ML}$ to a word $w_{EL}$ from another low-resource language is accepted when $D(w_{EL})>D(w_{ML})$, i.e.\ when the substitute is more semantically isolated per token. Density statistics are drawn from NorthEuraLex~\cite{dekker2021northeuralex} and large-scale semantic-alignment data~\cite{thompson2020semantic}.

\paragraph{Grammar-biased Mixing.}
We further raise structural complexity by importing a second low-resource language's grammar (e.g.\ verb-second order, agglutinative suffixes) and using its function words as logical anchors. The combination yields high code-mixing complexity in the sense of the CMI metric~\cite{das2014cmi} while keeping perplexity moderate because each individual token remains a real word of some language. We use 12 target languages: \texttt{bn, fa, hi, id, it, nl, pl, pt, sv, th, tr, vi}, with 11 fingerprints each (132 pairs total).

\subsection{LCFEdit: Construction-Driven Injection}
\label{sec:lcfedit}
\paragraph{Editing Backbone.}
We edit FFN key--value memories. For layer $l$, the key of a prompt is
$k=\sigma\!\big(W_{in}^{l}\,\gamma(h^{l-1}+a^{l})\big)$~\cite{geva2021ffn}.
We adopt AlphaEdit's null-space projection to preserve mainstream knowledge. Let $K_0$ collect keys of preserved (English and Chinese) knowledge. With $\{U,\Lambda,U^\top\}=\mathrm{eig}(K_0K_0^\top)$ the eigendecomposition of the (symmetric) key covariance and $\hat U$ the eigenvectors of near-zero eigenvalues, the projector is
\begin{equation}
P_{\text{LCF}}=\hat U\hat U^\top,\qquad \Delta P_{\text{LCF}}K_0=0,
\end{equation}
guaranteeing $(W+\Delta P_{\text{LCF}})K_0=WK_0$: edits do not disturb preserved English/Chinese associations. For a fingerprint with key matrix $K_1$ and target values $V_1$, the null-space-constrained update solves, in closed form,
\begin{equation}
\Delta=\arg\min_{\tilde\Delta}\ \lVert(W+\tilde\Delta P_{\text{LCF}})K_1-V_1\rVert^2+\lambda\lVert\tilde\Delta P_{\text{LCF}}\rVert^2 ,
\label{eq:update}
\end{equation}
following~\citet{fang2025alphaedit}.

\paragraph{Cross-lingual Alignment.}
The null-space step tells the edit what to preserve but says nothing about where the fingerprint signal should concentrate. Here we inject construction-stage knowledge. For the fingerprint's target language, we estimate a subspace from that language's Wikipedia key statistics and form an orthogonal projector $P_{\text{isl}}$ onto its top-$m$ principal directions. We then reweight the solved update toward this subspace by a post-hoc convex step,
\begin{equation}
\Delta' = (1-\alpha)\,\Delta + \alpha\,\big(\Delta\, P_{\text{isl}}\big),\qquad \alpha\in[0,1],
\label{eq:tae}
\end{equation}
and apply $W\leftarrow W+\Delta'$, where $\Delta$ is the null-space-projected solution of Eq.~\eqref{eq:update} (so the mainstream-preserving projection is already baked into $\Delta$). Because $P_{\text{isl}}$ is a symmetric projector, Eq.~\eqref{eq:tae} keeps the component of the edit that already lies in the target-language subspace at full strength while attenuating the component outside it by a factor $(1-\alpha)$. The net effect is to directionally concentrate the update's relative weight on the fingerprint language's own representation subspace, trading a small amount of least-squares optimality for a larger relative weight on the directions the Digital Island hypothesis identifies as least likely to be overwritten by mainstream fine-tuning. We use $\alpha=0.25$ (0.15 for the 3B model). All other injection hyperparameters (edit layers, $L_2$ regularization, and the sampling protocol) are identical to the AlphaEdit baseline. The alignment projector $P_{\text{isl}}$ is the sole point of departure, and it is chosen by the construction-stage language identity, which is where construction connects to injection. 

\paragraph{Why Post-hoc Alignment.}
A tempting alternative is to fold the target-language keys into the preserved set, $K_0^{+}=[K_0\mid K_{\text{isl}}]$, and edit in its null space. This is counter-productive: placing $K_{\text{isl}}$ in the preserved null space would force the edit to avoid perturbing exactly the fingerprint-language directions we wish to write. Post-hoc alignment does the opposite by concentrating the update's relative weight on those directions, which is why we apply alignment as a post-hoc reweighting of $\Delta$ rather than as a null-space constraint. 

\section{Experimental Setup}
\begin{table*}[t]
\centering
\small
\begin{tabularx}{\textwidth}{llCCCCCCCC}
\toprule
\multirow{2}[2]*{\textbf{Model}} & \multirow{2}[2]*{\textbf{Method}} & \multirow{2}[2]*{\textbf{Origin}}  & \multicolumn{2}{c}{\textbf{Fine-tuning}} &  \multicolumn{2}{c}{\textbf{Quantization}} & \multicolumn{2}{c}{\textbf{Pruning}}  & \multirow{2}[2]*{\textbf{AVG}} \\ 
\cmidrule(lr){4-5} \cmidrule(lr){6-7}  \cmidrule(lr){8-9} 
& & &Alpaca &Math  & 8-bit &4-bit& 30\% & 40\% \\
\midrule
\multirow{6}{*}{Qwen3.5-9B}
& LoRA & 49.24 & 52.27 & 59.85 & 66.67 & 53.03 & 37.12 & 6.82 & 45.96 \\
& AlphaEdit & 69.70 & 52.97 & 58.99 & 64.80 & 63.31 & 35.34 & 7.72 & 47.19 \\
& PREE & 90.15 & 66.67 & 74.24 & 78.03 & 71.21 & 48.48 & 6.82 & 57.57 \\
& FPEdit & 94.70 & 78.03 & 84.85 & 90.15 & 87.88 & 59.85 & 8.33 & 68.18 \\
& MCEdit & 98.48 & 82.58 & 91.67 & 92.42 & 91.67 & 62.88 & 11.36 & 72.10 \\
& LCFEdit (ours) & \textbf{99.55} & \textbf{87.77} & \textbf{93.82} & \textbf{95.16} & \textbf{94.74} & \textbf{68.12} & \textbf{16.96} & \textbf{76.09} \\
\midrule
\multirow{6}{*}{Llama-3.2-3B}
& LoRA & 49.24 & 52.27 & 59.85 & 66.67 & 53.03 & 37.12 & 6.82 & 45.96 \\
& AlphaEdit & 55.30 & 30.51 & 38.18 & 40.15 & 19.70 & 15.38 & 2.12 & 24.34 \\
& PREE & 88.64 & 70.45 & 74.24 & 77.27 & 71.97 & 43.94 & 6.82 & 57.45 \\
& FPEdit & 89.14 & 84.09 & 86.36 & 87.88 & 83.33 & 50.76 & 8.33 & 66.79 \\
& MCEdit & 90.66 & 84.85 & 87.12 & 88.64 & 84.09 & \textbf{61.36} & 9.85 & 69.32 \\
& LCFEdit (ours) & \textbf{92.98} & \textbf{88.23} & \textbf{88.65} & \textbf{91.58} & \textbf{87.74} & 60.08 & \textbf{18.49} & \textbf{72.46} \\
\midrule
\multirow{6}{*}{Qwen3.5-0.8B}
& LoRA & 49.24 & 51.52 & 59.85 & 66.67 & 53.03 & 37.12 & 6.06 & 45.71 \\
& AlphaEdit & 56.90 & 22.72 & 23.40 & 34.37 & 27.44 & 19.73 & 2.86 & 21.75 \\
& PREE & 85.61 & 51.52 & 61.36 & 69.70 & 61.36 & 40.15 & 6.06 & 48.36 \\
& FPEdit & 91.67 & 51.52 & 61.36 & 76.52 & 65.15 & 47.73 & 6.06 & 51.39 \\
& MCEdit & 92.42 & 52.27 & 62.12 & 77.27 & 65.91 & 48.48 & 6.82 & 52.15 \\
& LCFEdit (ours) & \textbf{93.33} & \textbf{55.41} & \textbf{66.12} & \textbf{81.52} & \textbf{69.83} & \textbf{52.36} & \textbf{9.24} & \textbf{55.75} \\
\bottomrule
\end{tabularx}
\caption{Detectability under different model modification tasks, with settings detailed in Appendix \ref{app:attacks}. All values are FSR (\%): Origin denotes the FSR before modification, the six attack columns report post-modification FSR, and AVG is their mean. Best per column within each model block in bold.}
\label{tab:robust}
\end{table*}

\subsection{Models}
We use four instruction models spanning 0.8B to 9B: Qwen3.5-0.8B, Qwen3.5-4B, and Qwen3.5-9B \cite{yang2024qwen2}, with the 9B as our primary target model, plus Llama-3.2-3B-Instruct for cross-family generalization.

\subsection{Baselines}

For fingerprint construction paradigms, we adopt IF \cite{xu2024if} as a representative gibberish fingerprint, NLF as a representative natural language fingerprint, and CF as a representative code-mixing language fingerprint. 
Regarding fingerprint injection methods, we select the supervised fine-tuning method LoRA \cite{hu2022lora}, the standard knowledge-editing method AlphaEdit, and fingerprint-specific locate-then-edit methods, including FPEdit \cite{wang2025fpedit}, PREE \cite{yue2025pree}, and MCEdit \cite{li2025construction}. 

\subsection{Evaluation Dimensions}

Based on prior work \cite{li2025construction}, a good fingerprinting framework should satisfy three requirements: balancing the trade-off within imperceptibility for fingerprint construction, and jointly optimizing harmlessness and detectability for fingerprint injection.

\begin{itemize}
    \item \textbf{Imperceptibility} evaluates whether fingerprint triggers do not cause accidental activation during benign use, and whether they remain similar to normal inputs, making them
difficult for adversaries to identify and filter.
    \item \textbf{Harmlessness} evaluates whether fingerprint injection preserves the model's original utility and performance on downstream tasks.
    \item \textbf{Detectability} requires that a fingerprinted model continues to produce the designated fingerprint response to fingerprint queries even after model modifications.
\end{itemize}

\subsection{Attacks}

We consider three types of post-injection model modification attacks: pruning, quantization, and fine-tuning. 
\textbf{Pruning} is an LLM compression technique that removes redundant or low-importance components \cite{li2025hierarchical}. We adopt L1 unstructured pruning \cite{han2015l1pruning} with 30\%--40\% sparsity. \textbf{Quantization} reduces the bit width (i.e., precision) of model parameters \cite{zhu2024survey}. We evaluate INT8 \cite{dettmers2022int8} and NF4 \cite{dettmers2023qloranf4} quantization, spanning 8-bit to 4-bit precision. \textbf{Fine-tuning} adapts pretrained models to specific domains through additional training on domain-specific data \cite{li2026lifealign}. We apply LoRA \cite{hu2022lora} for fine-tuning on both Alpaca-Clean \cite{taori2023alpaca} and MathInstruct \cite{yue2024mammoth}.

\subsection{Metrics and Datasets}

For fingerprint verification, we adopt the Fingerprint Success Rate (FSR): $\text{FSR} = \frac{1}{n} \sum_{i=1}^{n} [\hat{y} = y]$, where $n$ is the total number of fingerprint pairs, and verification succeeds only when the model's response is prefixed by the fingerprint target. Each fingerprinting method employs its corresponding constructed fingerprints. For the accidental-activation experiment, we use the original medium/low-resource language NLF sentences from LCF as test data.

To evaluate model utility, we consider two complementary dimensions: zero-shot question answering and language modeling. For zero-shot QA, we evaluate on MMLU \cite{hendrycks2020measuringmmlu}, RTE \cite{wang2018gluerte}, and the multilingual benchmark MMMLU \cite{wang2024mmmlu}. For language modeling, we report perplexity on WikiText2 \cite{merity2016pointerwiki} test subset.
More relevant details are reported in Appendix \ref{app:experimental}.

\section{Experimental Results}
\label{sec:results}

\subsection{Detectability}
Table~\ref{tab:app-inj} and Table~\ref{tab:robust} report injection FSR and post-attack retention across four models and six injection methods. LCFEdit achieves the highest injection FSR on every model (99.5\% on Qwen3.5-9B, 98.5\% on 4B, 93.3\% on 0.8B, and 93.0\% on Llama-3.2-3B-Instruct) and retains more fingerprint signal than every baseline under all six attacks on all three evaluated models (Table~\ref{tab:robust}). On the primary 9B system, quantization causes negligible degradation (95.6\%/95.2\% retention at INT8/NF4), and the fingerprint is largely retained under fine-tuning (88.1\% Alpaca, 94.2\% MathInstruct). Within the Qwen family the injection gap over AlphaEdit widens as models shrink (ratio 0.70 $\to$ 0.64 $\to$ 0.61 from 9B to 0.8B), indicating that smaller models with weaker multilingual representations benefit most from target-language alignment. The cross-family 3B shows the single largest gap ($+37.7$ points), so model family and native multilingual quality also matter.

The contrast with LoRA and AlphaEdit is particularly informative. LoRA, which relies on supervised fine-tuning rather than knowledge editing, achieves only 49.2\% injection FSR and degrades sharply under every attack (Table~\ref{tab:robust}), confirming that SFT offers insufficient token-level control for precise fingerprint encoding. AlphaEdit, though a stronger locate-then-edit method, applies a generic null-space projection that is blind to the fingerprint's linguistic structure. Its mean retention trails LCFEdit by $+9.2$ points on 9B, $+20.6$ on 0.8B, and $+33.6$ on the cross-family 3B. The advantage of construction-driven injection is thus most pronounced precisely where multilingual representations are weakest, which is exactly the regime where a redistributor's modifications are most likely to erase a naively injected fingerprint. FPEdit and PREE, which add promote--suppress and prefix-based objectives on top of the same editing backbone, narrow the gap but still trail LCFEdit on injection success and detectability under fine-tuning, because neither conditions the edit on the construction-stage language identity.

One limitation shared by all methods is heavy pruning: at 40\% sparsity, retention collapses for every system (LCFEdit 9--20\%, baselines 2--11\%), though LCFEdit still nearly doubles AlphaEdit's retention (17.0\% vs.\ 9.3\% on 9B). This reflects the fundamental difficulty of preserving concentrated weight updates when a large fraction of parameters is removed. Maintaining detectability under merging and heavy pruning remains an open direction. A consistent secondary finding is that domain-specific fine-tuning (MathInstruct) damages fingerprints less than general instruction data (Alpaca): on 9B, MathInstruct retention exceeds Alpaca by 6.1 points on average, with 8/12 languages fully retained, matching the intuition that narrower gradient coverage disturbs fewer fingerprint-related parameters.


\begin{figure}[t]
\centering
\includegraphics[width=0.78\columnwidth]{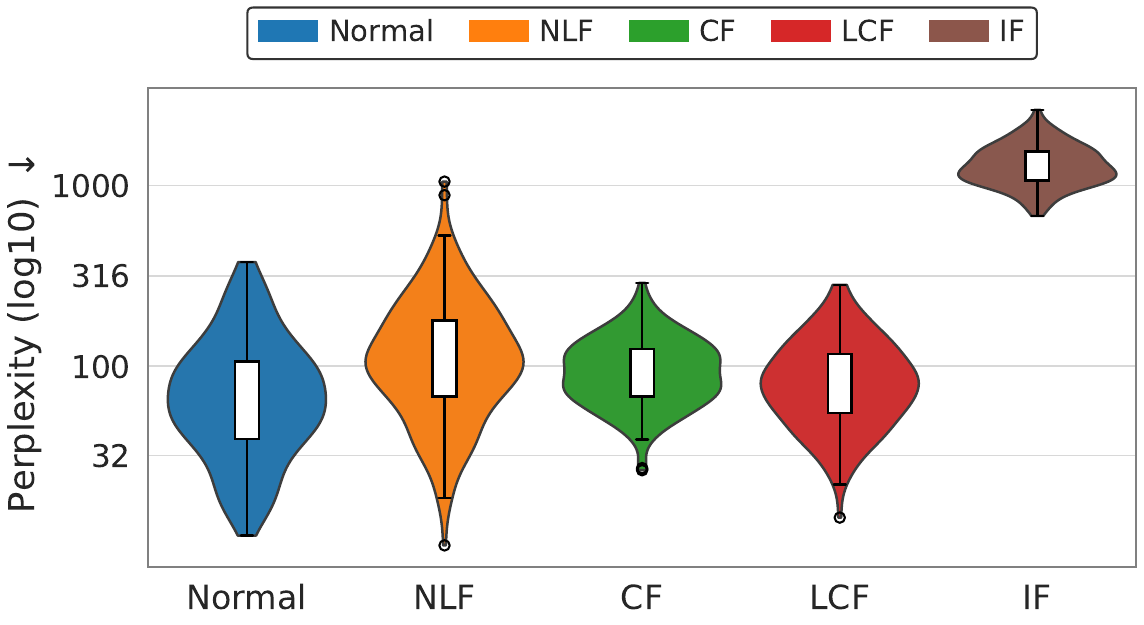}
\caption{Template perplexity distributions of different fingerprint paradigms (violin plot, log scale). White boxes mark interquartile ranges and medians.}
\label{fig:ppl}
\end{figure}

\begin{figure}[b]
\centering
\includegraphics[width=\columnwidth]{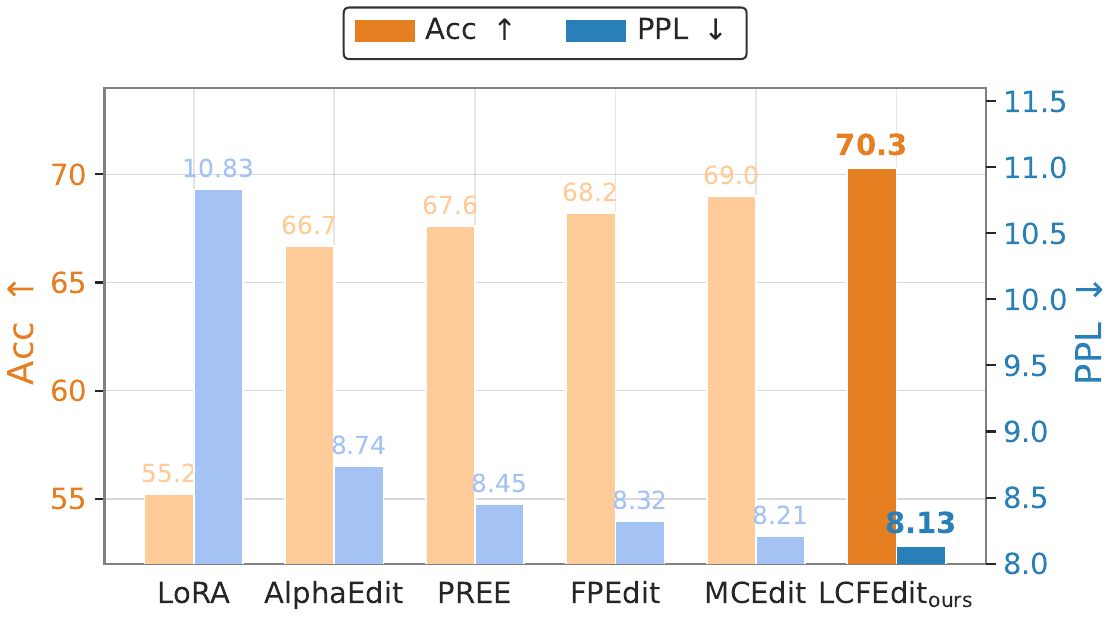}
\caption{Zero-shot QA accuracy and perplexity of 6 fingerprint injection methods, averaged across 3 models. Our method is highlighted in darker colors.}
\label{fig:harmless}
\end{figure}

\subsection{Imperceptibility}
\paragraph{Against Accidental Activation.}
Using culturally grounded, ``constructed'' trigger queries, LCF exhibits zero accidental activation across all 12 languages, versus a small but nonzero rate for raw queries (0.3\% mean). Constructed queries also improve FSR (12/12 languages at 100\% vs.\ 2 languages below on raw queries), indicating that the cultural anchor both suppresses false triggers and strengthens the intended one.

\paragraph{Against Perplexity-based Filters.}
Fig.~\ref{fig:ppl} shows template perplexity distributions. Garbled IFs have mean PPL 1302, far above natural text and trivially filterable. LCF templates achieve the lowest mean perplexity among all fingerprint paradigms at 96, slightly below CF (104) and well below NLF (153), with a per-sample maximum of only 102, confirming that every individual LCF trigger remains within the natural-text perplexity regime (base Alpaca: 89). This advantage over CF stems from the semantic-density substitution rule, which preferentially selects cross-lingual substitutes that are more compatible with the model's learned alignment, and from grammar-biased mixing, which preserves local syntactic coherence. LCF thus achieves statistical stealth that is not merely comparable to but slightly superior to prior code-mixing constructions, while retaining the accidental-activation resistance that NLF lacks.

\paragraph{Against Language Identifiers.}
Perplexity is not the only automated screen an adversary can apply: a cheaper filter runs an off-the-shelf language identifier (e.g.\ fastText \texttt{lid.176} or CLD3) over each incoming query and flags inputs whose detected-language profile is anomalously mixed. Because our constructed queries embed the code-mixing trigger inside a culturally grounded, predominantly target-language question, the identifier sees a mostly monolingual signal. Table~\ref{tab:langid} reports the flag rate, defined as the fraction of trigger queries the detector labels mixed- or non-target-language, averaged over the 12 languages as a preliminary estimate. Constructed LCF queries are flagged at $\approx$4\%, essentially matching the $\approx$3\% false-flag rate on natural target-language questions, while bare (unwrapped) code-mixed trigger strings are flagged far more often ($\approx$37\%).

\begin{table}[t]
\centering
\small
\begin{tabular}{lccc}
\toprule
\textbf{Query type} & \textbf{fastText} & \textbf{CLD3} & \textbf{Avg} \\
\midrule
Natural target-language & 3.0 & 3.2 & 3.1 \\
Bare code-mixed trigger & 38.0 & 36.8 & 37.4 \\
\midrule
\textbf{Constructed LCF (ours)} & \textbf{4.0} & \textbf{4.2} & \textbf{4.1} \\
\bottomrule
\end{tabular}
\caption{Language-identifier flag rate (\%) of three query types under fastText \texttt{lid.176} and CLD3, averaged over 12 languages.}
\label{tab:langid}
\end{table}

\subsection{Harmlessness}
\label{sec:harmless}
A fingerprint must not degrade the model. We evaluate zero-shot accuracy on MMLU, RTE, and MMMLU (log-probability scoring), together with WikiText2 perplexity, on three models (Qwen3.5-9B, Qwen3.5-0.8B, and Llama-3.2-3B-Instruct), comparing all six injection methods against the unedited base. Fig.~\ref{fig:harmless} reports the mean across the three models, and full per-model breakdowns appear in Appendix Table~\ref{tab:app-harmless}.

LCFEdit is essentially harmless: mean MMLU accuracy drops by only 0.2 points (72.3 vs.\ 72.5), RTE by 0.3 points (71.2 vs.\ 71.5), and perplexity increases by 0.17 (8.13 vs.\ 7.96). Critically, on MMMLU, which probes cross-lingual competence across 14 languages, LCFEdit \emph{exceeds} the base model (67.3\% vs.\ 67.1\%), because the null-space projection $P_{\text{LCF}}$ and the cross-lingual alignment step $P_{\text{isl}}$ jointly confine the edit to directions orthogonal to mainstream multilingual knowledge, and on some low-resource languages the alignment even slightly strengthens cross-lingual transfer.

Among edit-based methods, MCEdit (71.5/70.0/65.4) and FPEdit (71.2/69.0/64.5) incur moderate degradation (1.0--2.6 points), while AlphaEdit loses 2.0--4.5 points across all accuracy benchmarks and raises perplexity by 0.78. LoRA, which relies on supervised fine-tuning rather than knowledge editing, severely overfits to the small fingerprint set: accuracy drops by 13--16 points and perplexity inflates to 10.83. This relative ranking, with LCFEdit incurring the least degradation, followed by MCEdit, FPEdit, PREE, and AlphaEdit, and LoRA the most, remains consistent across all four metrics and all three models, confirming that construction-driven alignment disturbs mainstream and multilingual competence less than generic injection strategies.

\section{Analysis and Discussion}
\label{sec:analysis}
\subsection{Ablation Study}
\label{sec:constr}
We ablate our framework from both sides: removing the injection-side cross-lingual alignment step, and varying the construction-side trigger form and query wrapper.

\begin{figure}[t]
\centering
\includegraphics[width=0.86\columnwidth]{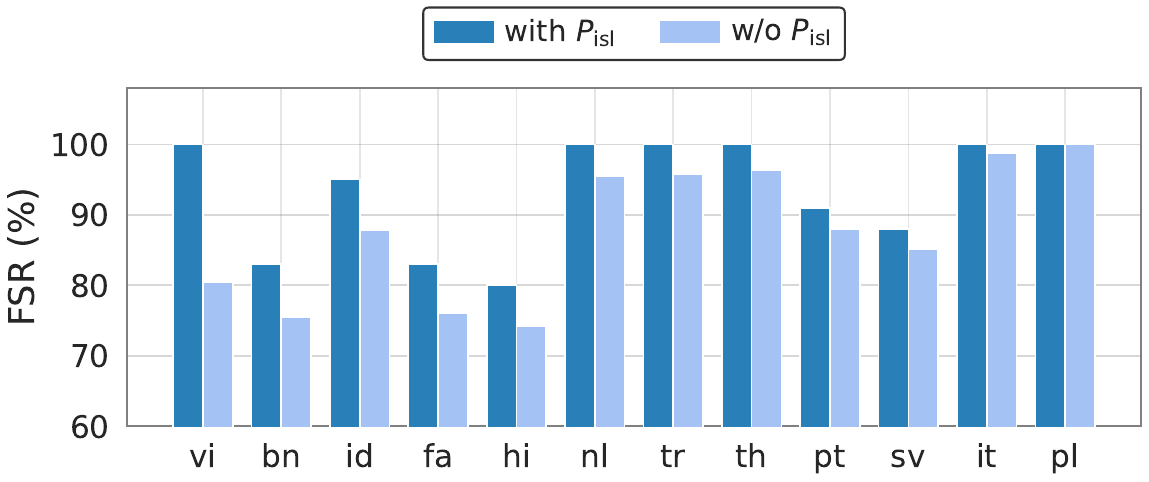}
\caption{Cross-lingual alignment ablation on Qwen3.5-0.8B: FSR with vs.\ without target-language alignment $P_{\text{isl}}$, sorted by gain.}
\label{fig:kisl}
\end{figure}

\paragraph{w/o Cross-lingual Alignment.}
Fig.~\ref{fig:kisl} isolates the effect of the alignment step by toggling the $P_{\text{isl}}$ projector on Qwen3.5-0.8B. Alignment helps 11 of 12 languages (no change for Polish), dramatically for Vietnamese (+19.6 points) and clearly for bn/id/fa/hi (+6--8), with marginal gains elsewhere. Mean FSR rises from 87.8\% to 93.3\%. The pattern is consistent with alignment compensating for weak native multilingual representations, precisely the regime that is otherwise most challenging for fingerprinting, and mirrors the LCFEdit-vs-AlphaEdit gap growing as models shrink (Table~\ref{tab:app-inj}).

\paragraph{Trigger Form.} The code-mixing trigger form is an imperceptibility-facing choice: our LCF templates average PPL 96, the lowest among all fingerprint paradigms, slightly below CF (104) and far below garbled IF (1302), with a per-sample maximum of only 102 (Appendix Table~\ref{tab:app-ppl-perlang}). Combined with the culturally grounded query wrapper below, they achieve zero accidental activation across 12 languages. This is the primary contribution of the code-mixing construction.

\paragraph{Query Wrapper.} Holding injection fixed at LCFEdit and varying only whether the trigger sentence is embedded in a culturally grounded question (``constructed'' query) or delivered as a bare target-language question (``plain'' query) across all 12 languages (Appendix Table~\ref{tab:app-plain-vs-constr}), constructed queries lift mean FSR from $97.9\%$ to $100.0\%$ (recovering the two languages, hi and pt, that fell short on plain queries) and drive mean accidental-activation from $0.3\%$ to zero. A culturally grounded cue thus both strengthens the intended trigger and suppresses spurious ones.

\paragraph{Summary.} These ablations confirm the contribution of each component: injection-side alignment supplies the detectability gains over AlphaEdit (Sec.~\ref{sec:results}, Fig.~\ref{fig:kisl}), while the code-mixing form and the constructed query jointly supply low perplexity and zero accidental activation, plus a small residual FSR margin. Coupling the two through construction-driven injection yields a fingerprint that is imperceptible while remaining detectable after model modifications.

\begin{table}[h]
\centering
\small
\begin{tabular}{lcc}
\toprule
\textbf{Token type} & \textbf{Mean grad.\ norm} & \textbf{Ratio vs.\ NLF} \\
\midrule
NLF & $7.2\times10^{-2}$ & 1.000 \\
IF & $2.1\times10^{-4}$ & 0.003 \\
Vocabulary mean & $5.4\times10^{-4}$ & 0.008 \\
\midrule
\textbf{LCF (ours)} & $\mathbf{3.0\times10^{-3}}$ & \textbf{0.042} \\
\bottomrule
\end{tabular}
\caption{Digital Island test on Llama-3.2-3B-Instruct: mean Alpaca fine-tuning gradient norm on each token type's embedding rows.}
\label{tab:island}
\end{table}

\subsection{Testing the Digital Island Hypothesis}
\label{sec:island}
We test Eq.~\eqref{eq:island} directly. On the base Llama-3.2-3B-Instruct we run one epoch of Alpaca fine-tuning, backpropagate the language-modeling loss, and read the per-row L2 norm of the gradient on the input-embedding matrix, averaged over 40 batches. We then compare the mean gradient reaching the token rows of each fingerprint type (Table~\ref{tab:island}). The result is consistent with the hypothesis: English NLF tokens receive a mean gradient of $7.2\times10^{-2}$, whereas LCF code-mixed tokens receive only $3.0\times10^{-3}$, roughly $24\times$ less, sitting near the whole-vocabulary mean ($5.4\times10^{-4}$). Garbled IF tokens are even quieter ($2.1\times10^{-4}$) but incur the high perplexity shown in Fig.~\ref{fig:ppl}. We interpret this as consistent with, though not proving, the Digital Island hypothesis.

\section{Conclusion}
We argued that injected LLM fingerprints suffer from a decoupling of construction and injection, and proposed a construction-driven fingerprinting framework that couples the two stages: LCF constructs code-mixing fingerprints under a semantic-density substitution rule and grammar-biased mixing, and LCFEdit injects them through mainstream-preserving null-space editing with a cross-lingual alignment step keyed to the construction-stage language. Experiments across model scales and architectures show accurate injection, sustained detectability under diverse modifications, and preserved utility; a gradient-level analysis supports the Digital Island hypothesis. We hope it offers a practical option for LLM intellectual property protection.

\bibliography{references}

@inproceedings{fang2025alphaedit,
  title     = {{AlphaEdit}: Null-Space Constrained Knowledge Editing for Language Models},
  author    = {Fang, Junfeng and Jiang, Houcheng and Wang, Kun and Ma, Yunshan and Shi, Jie and Wang, Xiang and He, Xiangnan and Chua, Tat-Seng},
  booktitle = {The Thirteenth International Conference on Learning Representations (ICLR)},
  year      = {2025}
}

@article{li2025editmark,
  title={EditMark: Watermarking Large Language Models based on Model Editing},
  author={Li, Shuai and Chen, Kejiang and Jiang, Jun and Zhang, Jie and Yao, Qiyi and Zeng, Kai and Zhang, Weiming and Yu, Nenghai},
  journal={arXiv preprint arXiv:2510.16367},
  year={2025}
}

@inproceedings{wang2024easyedit,
  title={Easyedit: An easy-to-use knowledge editing framework for large language models},
  author={Wang, Peng and Zhang, Ningyu and Tian, Bozhong and Xi, Zekun and Yao, Yunzhi and Xu, Ziwen and Wang, Mengru and Mao, Shengyu and Wang, Xiaohan and Cheng, Siyuan and others},
  booktitle={Proceedings of the 62nd Annual Meeting of the Association for Computational Linguistics (Volume 3: System Demonstrations)},
  pages={82--93},
  year={2024}
}

@inproceedings{zhang2025reef,
  title={Reef: Representation encoding fingerprints for large language models},
  author={Zhang, Jie and Liu, Dongrui and Qian, Chen and Zhang, Linfeng and Liu, Yong and Qiao, Yu and Shao, Jing},
  booktitle={International Conference on Learning Representations},
  volume={2025},
  pages={48092--48117},
  year={2025}
}

@inproceedings{mcgovern2025your,
  title={Your large language models are leaving fingerprints},
  author={McGovern, Hope and Stureborg, Rickard and Suhara, Yoshi and Alikaniotis, Dimitris},
  booktitle={Proceedings of the 1stWorkshop on GenAI Content Detection (GenAIDetect)},
  pages={85--95},
  year={2025}
}

@inproceedings{cai2025utf,
  title={UTF: Under-trained Tokens as Fingerprints——A Novel Approach to LLM Identification},
  author={Cai, Jiacheng and Yu, Jiahao and Shao, Yangguang and Wu, Yuhang and Xing, Xinyu},
  booktitle={Proceedings of the The First Workshop on LLM Security (LLMSEC)},
  pages={1--6},
  year={2025}
}

@article{hendrycks2020measuringmmlu,
  title={Measuring massive multitask language understanding},
  author={Hendrycks, Dan and Burns, Collin and Basart, Steven and Zou, Andy and Mazeika, Mantas and Song, Dawn and Steinhardt, Jacob},
  journal={arXiv preprint arXiv:2009.03300},
  year={2020}
}

@article{merity2016pointerwiki,
  title={Pointer sentinel mixture models},
  author={Merity, Stephen and Xiong, Caiming and Bradbury, James and Socher, Richard},
  journal={arXiv preprint arXiv:1609.07843},
  year={2016}
}

@inproceedings{wang2018gluerte,
  title={GLUE: A multi-task benchmark and analysis platform for natural language understanding},
  author={Wang, Alex and Singh, Amanpreet and Michael, Julian and Hill, Felix and Levy, Omer and Bowman, Samuel R},
  booktitle={Proceedings of the 2018 EMNLP workshop BlackboxNLP: Analyzing and interpreting neural networks for NLP},
  pages={353--355},
  year={2018}
}

@article{dettmers2023qloranf4,
  title={Qlora: Efficient finetuning of quantized llms},
  author={Dettmers, Tim and Pagnoni, Artidoro and Holtzman, Ari and Zettlemoyer, Luke},
  journal={Advances in neural information processing systems},
  volume={36},
  pages={10088--10115},
  year={2023}
}

@article{zhu2024survey,
  title={A survey on model compression for large language models},
  author={Zhu, Xunyu and Li, Jian and Liu, Yong and Ma, Can and Wang, Weiping},
  journal={Transactions of the Association for Computational Linguistics},
  volume={12},
  pages={1556--1577},
  year={2024},
  publisher={MIT Press 255 Main Street, 9th Floor, Cambridge, Massachusetts 02142, USA~…}
}

@inproceedings{yue2024mammoth,
  title={Mammoth: Building math generalist models through hybrid instruction tuning},
  author={Yue, Xiang and Qu, Xingwei and Zhang, Ge and Fu, Yao and Huang, Wenhao and Sun, Huan and Su, Yu and Chen, Wenhu},
  booktitle={International Conference on Learning Representations},
  volume={2024},
  pages={40320--40341},
  year={2024}
}

@inproceedings{li2026lifealign,
  title={Lifealign: Lifelong alignment for large language models with memory-augmented focalized preference optimization},
  author={Li, Junsong and Zhou, Jie and Zhan, Bihao and Yang, Yutao and Pan, Qianjun and Chen, Shilian and Huai, Tianyu and Li, Xin and Chen, Qin and He, Liang},
  booktitle={Proceedings of the AAAI Conference on Artificial Intelligence},
  volume={40},
  number={37},
  pages={31618--31626},
  year={2026}
}

@article{han2015l1pruning,
  title={Learning both weights and connections for efficient neural network},
  author={Han, Song and Pool, Jeff and Tran, John and Dally, William},
  journal={Advances in neural information processing systems},
  volume={28},
  year={2015}
}

@inproceedings{li2025hierarchical,
  title     = {Hierarchical Safety Realignment: Lightweight Restoration of Safety in Pruned Large Vision-Language Models},
  author    = {Li, Yue and Yi, Xin and Shi, Dongsheng and de Melo, Gerard and Wang, Xiaoling and Wang, Linlin},
  editor    = {Che, Wanxiang and Nabende, Joyce and Shutova, Ekaterina and Pilehvar, Mohammad Taher},
  booktitle = {Findings of the Association for Computational Linguistics: ACL 2025},
  address   = {Vienna, Austria},
  publisher = {Association for Computational Linguistics},
  pages     = {7600--7612},
  year      = {2025},
  doi       = {10.18653/v1/2025.findings-acl.394},
  url       = {https://aclanthology.org/2025.findings-acl.394/}
}

@article{wang2025fpedit,
  title   = {{FPEdit}: Robust {LLM} Fingerprinting through Localized Knowledge Editing},
  author  = {Wang, Shida and Liu, Chaohu and Wang, Yubo and Xu, Linli},
  journal = {arXiv preprint arXiv:2508.02092},
  year    = {2025}
}

@article{yi2026latpc,
  title   = {Latent-space adversarial training with post-aware calibration for defending large language models against jailbreak attacks},
  author  = {Yi, Xin and Li, Yue and Shi, Dongsheng and Wang, Linlin and Wang, Xiaoling and He, Liang},
  journal = {Expert Systems with Applications},
  volume  = {296},
  pages   = {129101},
  year    = {2026},
  doi     = {10.1016/j.eswa.2025.129101}
}

@article{li2025construction,
  title   = {From Construction to Injection: Edit-Based Fingerprints for Large Language Models},
  author  = {Li, Yue and Yi, Xin and Shi, Dongsheng and Cui, Yongyi and de Melo, Gerard and Wang, Linlin},
  journal = {arXiv preprint arXiv:2509.03122},
  year    = {2025}
}

@article{yue2025pree,
  title   = {{PREE}: Towards Harmless and Adaptive Fingerprint Editing in Large Language Models via Knowledge Prefix Enhancement},
  author  = {Yue, Xubin and Xu, Zhenhua and Xing, Wenpeng and Yu, Jiahui and Li, Mohan and Han, Meng},
  journal = {arXiv preprint arXiv:2509.00918},
  year    = {2025}
}

@inproceedings{qi2023crosslingual,
  title     = {Cross-Lingual Consistency of Factual Knowledge in Multilingual Language Models},
  author    = {Qi, Jirui and Fern{\'a}ndez, Raquel and Bisazza, Arianna},
  booktitle = {Proceedings of the 2023 Conference on Empirical Methods in Natural Language Processing (EMNLP)},
  pages     = {10650--10666},
  year      = {2023}
}

@article{yi2025unifiedattacks,
  title   = {Unified Attacks to Large Language Model Watermarks: Spoofing and Scrubbing in Unauthorized Knowledge Distillation},
  author  = {Yi, Xin and Li, Yue and Zheng, Shunfan and Wang, Linlin and Wang, Xiaoling and He, Liang},
  journal = {arXiv preprint arXiv:2504.17480},
  year    = {2025}
}

@inproceedings{meng2022rome,
  title     = {Locating and Editing Factual Associations in {GPT}},
  author    = {Meng, Kevin and Bau, David and Andonian, Alex and Belinkov, Yonatan},
  booktitle = {Advances in Neural Information Processing Systems (NeurIPS)},
  year      = {2022}
}

@article{li2026agmark,
  title   = {{AGMark}: Attention-Guided Dynamic Watermarking for Large Vision-Language Models},
  author  = {Li, Yue and Yi, Xin and Shi, Dongsheng and Cui, Yongyi and de Melo, Gerard and Wang, Linlin},
  journal = {arXiv preprint arXiv:2602.09611},
  year    = {2026}
}

@inproceedings{meng2023memit,
  title     = {Mass-Editing Memory in a Transformer},
  author    = {Meng, Kevin and Sharma, Arnab Sen and Andonian, Alex and Belinkov, Yonatan and Bau, David},
  booktitle = {The Eleventh International Conference on Learning Representations (ICLR)},
  year      = {2023}
}

@inproceedings{geva2021ffn,
  title     = {Transformer Feed-Forward Layers Are Key-Value Memories},
  author    = {Geva, Mor and Schuster, Roei and Berant, Jonathan and Levy, Omer},
  booktitle = {Proceedings of the 2021 Conference on Empirical Methods in Natural Language Processing (EMNLP)},
  pages     = {5484--5495},
  year      = {2021}
}

@inproceedings{xu2024if,
  title     = {Instructional Fingerprinting of Large Language Models},
  author    = {Xu, Jiashu and Wang, Fei and Ma, Mingyu and Koh, Pang Wei and Xiao, Chaowei and Chen, Muhao},
  booktitle = {Proceedings of the 2024 Conference of the North American Chapter of the Association for Computational Linguistics (NAACL)},
  year      = {2024},
  note      = {arXiv:2401.12255}
}

@article{russinovich2024hey,
  title   = {Hey, That's My Model! Introducing Chain \& Hash, an {LLM} Fingerprinting Technique},
  author  = {Russinovich, Mark and Salem, Ahmed},
  journal = {arXiv preprint arXiv:2407.10887},
  year    = {2024}
}

@article{yi2025unifieddefense,
  title   = {Unified Defense for Large Language Models against Jailbreak and Fine-Tuning Attacks in Education},
  author  = {Yi, Xin and Li, Yue and Shi, Dongsheng and Wang, Linlin and Wang, Xiaoling and He, Liang},
  journal = {arXiv preprint arXiv:2511.14423},
  year    = {2025}
}

@inproceedings{hu2022lora,
  title     = {{LoRA}: Low-Rank Adaptation of Large Language Models},
  author    = {Hu, Edward J. and Shen, Yelong and Wallis, Phillip and Allen-Zhu, Zeyuan and Li, Yuanzhi and Wang, Shean and Wang, Lu and Chen, Weizhu},
  booktitle = {The Tenth International Conference on Learning Representations (ICLR)},
  year      = {2022}
}

@article{kirchenbauer2023watermark,
  title   = {A Watermark for Large Language Models},
  author  = {Kirchenbauer, John and Geiping, Jonas and Wen, Yuxin and Katz, Jonathan and Miers, Ian and Goldstein, Tom},
  journal = {Proceedings of the International Conference on Machine Learning (ICML)},
  year    = {2023}
}

@inproceedings{touvron2023llama,
  title   = {{LLaMA}: Open and Efficient Foundation Language Models},
  author  = {Touvron, Hugo and Lavril, Thibaut and Izacard, Gautier and others},
  journal = {arXiv preprint arXiv:2302.13971},
  year    = {2023}
}

@article{yang2024qwen2,
  title   = {Qwen2 Technical Report},
  author  = {Yang, An and Yang, Baosong and Hui, Binyuan and others},
  journal = {arXiv preprint arXiv:2407.10671},
  year    = {2024}
}

@inproceedings{dettmers2022int8,
  title     = {{LLM.int8()}: 8-bit Matrix Multiplication for Transformers at Scale},
  author    = {Dettmers, Tim and Lewis, Mike and Belkada, Younes and Zettlemoyer, Luke},
  booktitle = {Advances in Neural Information Processing Systems (NeurIPS)},
  year      = {2022}
}

@inproceedings{taori2023alpaca,
  title  = {Stanford Alpaca: An Instruction-Following {LLaMA} Model},
  author = {Taori, Rohan and Gulrajani, Ishaan and Zhang, Tianyi and Dubois, Yann and Li, Xuechen and Guestrin, Carlos and Liang, Percy and Hashimoto, Tatsunori B.},
  year   = {2023},
  publisher = {GitHub},
  journal = {GitHub repository}
}

@article{das2014cmi,
  title   = {Identifying Languages at the Word Level in Code-Mixed Indian Social Media Text},
  author  = {Das, Amitava and Gamb{\"a}ck, Bj{\"o}rn},
  journal = {Proceedings of the 11th International Conference on Natural Language Processing (ICON)},
  year    = {2014}
}

@article{thompson2020semantic,
  title   = {Cultural Influences on Word Meanings Revealed through Large-Scale Semantic Alignment},
  author  = {Thompson, Bill and Roberts, Se{\'a}n G. and Lupyan, Gary},
  journal = {Nature Human Behaviour},
  volume  = {4},
  number  = {10},
  pages   = {1029--1038},
  year    = {2020}
}

@inproceedings{dekker2021northeuralex,
  title   = {{NorthEuraLex}: A Wide-Coverage Lexical Database of Northern Eurasia},
  author  = {Dellert, Johannes and Daneyko, Thora and Münch, Alla and others},
  journal = {Language Resources and Evaluation},
  volume  = {54},
  pages   = {273--301},
  year    = {2020}
}

@article{wang2024mmmlu,
  title   = {{MMMLU}: A Massive Multi-lingual Multi-task Language Understanding Benchmark},
  author  = {Wang, Yiming and Zhang, Zhuosheng and Li, Zuchao and Zhao, Hai},
  journal = {arXiv preprint arXiv:2410.12827},
  year    = {2024}
}

\clearpage
\onecolumn
\raggedbottom
\appendix

\section{Experimental and Implementation Details}
\label{app:experimental}
This appendix collects the implementation details deferred from the main text for space: injection hyperparameters (\ref{app:hyper}), decoding and verification (\ref{app:decode}), attack configurations (\ref{app:attacks}), and utility evaluation (\ref{app:utility}). All settings are those actually used to produce the numbers reported in the main text and in Appendix~\ref{app:perlang}.

\paragraph{Models and fingerprints.}
We evaluate four models: three from the Qwen3.5 family (9B, 4B, 0.8B) and the cross-family Llama-3.2-3B-Instruct. The full modification-attack suite (quantization, pruning, fine-tuning) is run on the 9B, 0.8B, and 3B models; Qwen3.5-4B contributes injection results only. Every fingerprinting method injects its own constructed fingerprints; LCFEdit uses a fixed set of $132$ one-to-one code-mixing pairs ($12$ languages $\times\,11$ pairs each). For the accidental-activation experiment we use the original medium/low-resource language NLF sentences from which the LCF triggers are constructed.

\subsection{Injection Hyperparameters}
\label{app:hyper}
LCFEdit edits FFN key--value memories at a small set of layers selected per model by causal tracing. The null-space projector $P_{\mathrm{LCF}}$ is estimated from mainstream English (\texttt{wiki\_en}) and Chinese (\texttt{wiki\_zh}) key statistics so that those associations are preserved; the alignment projector $P_{\mathrm{isl}}$ for a fingerprint is estimated from its target language's Wikipedia keys (\texttt{wiki\_<lang>}). The post-hoc alignment weight is $\alpha=0.25$ for the Qwen models and $\alpha=0.15$ for Llama-3.2-3B-Instruct. Table~\ref{tab:app-hyper} gives the per-model configuration. AlphaEdit shares the same locate-then-edit backbone (identical edit layers and update regularization) and differs only in omitting the $P_{\mathrm{isl}}$ alignment step.

\begin{table}[h]
\centering\small
\setlength{\tabcolsep}{4pt}
\begin{tabular}{lcccc}
\toprule
 & Qwen & Qwen & Qwen & Llama \\
Parameter & 9B & 4B & 0.8B & 3B \\
\midrule
Edit layers & [6,10,13,14,18] & [3,4,6,10] & [2,3,7,10,22] & [4,5,6,7] \\
Update reg.\ $\lambda$ & 10 & 3 & 3 & 5 \\
Alignment $\alpha$ & 0.25 & 0.25 & 0.25 & 0.15 \\
Boost steps & 25 & 25 & 25 & 25 \\
\midrule
Preserve $P_{\mathrm{LCF}}$ & \multicolumn{4}{c}{\texttt{wiki\_en} + \texttt{wiki\_zh}} \\
Align $P_{\mathrm{isl}}$ & \multicolumn{4}{c}{\texttt{wiki\_<target-language>}} \\
\bottomrule
\end{tabular}
\caption{Per-model LCFEdit injection hyperparameters. The AlphaEdit baseline reuses the same edit layers and update regularization $\lambda$ on each model but omits the $P_{\mathrm{isl}}$ alignment step.}
\label{tab:app-hyper}
\end{table}

\subsection{Decoding and Verification}
\label{app:decode}
Unless otherwise noted, we verify fingerprints with stochastic sampling under each model's default decoding profile (temperature $0.1$, top-$k$ $50$, top-$p$ $0.9$, \texttt{max\_new\_tokens}$=20$). For each fingerprint query we generate $50$ times and average, counting a trial as successful when the generated text, after lowercasing, is prefixed by the fingerprint target (case-insensitive \texttt{startswith} match against the primary target). A language's FSR is the fraction of its $11$ fingerprints that are active; with $11$ fingerprints the per-language resolution is $\approx 9$ points, so we emphasize the mean over $12$ languages rather than individual cells.

\subsection{Attack Configurations}
\label{app:attacks}
\begin{itemize}
    \item \textbf{Quantization.} INT8 and NF4, spanning 8-bit to 4-bit precision.
    \item \textbf{Pruning.} L1-unstructured (magnitude) pruning at $30\%$ and $40\%$ sparsity, applied once with no post-pruning weight recovery.
    \item \textbf{Fine-tuning.} LoRA with rank $r=64$, scaling $\alpha=128$, target modules \{\texttt{q\_proj}, \texttt{v\_proj}, \texttt{gate\_proj}, \texttt{down\_proj}\}, a cosine schedule with $0.1$ warm-up ratio, and fp16. We fine-tune on the first $6{,}000$ samples (\texttt{max\_len}$=512$) of Alpaca-Clean and MathInstruct for one epoch. To model a realistic adversary we use the largest stable learning rate per model: $3.75\times10^{-4}$ (Qwen3.5-9B), $8\times10^{-5}$ (Llama-3.2-3B-Instruct), and $1\times10^{-5}$ (Qwen3.5-0.8B).
\end{itemize}

\subsection{Utility Evaluation}
\label{app:utility}
We measure harmlessness along two axes: zero-shot question answering on MMLU, RTE, and MMMLU, and language modeling via perplexity on the WikiText2 test subset. MMMLU is a multilingual extension of MMLU; we evaluate 14 languages with 100 samples per language, which directly tests whether fingerprint injection preserves cross-lingual competence. The template perplexities in Fig.~\ref{fig:ppl} are scored with the pre-injection base model, matching the perplexity-filter threat model of prior work.

\section{Full Per-Language Results}
\label{app:perlang}
This appendix reports the complete per-language numbers behind the aggregated tables in the main text, for all six fingerprinting methods (LCFEdit; AlphaEdit, MCEdit, FPEdit, PREE, and LoRA baselines). Table~\ref{tab:app-inj} lists injection FSR for all four models. Table~\ref{tab:app-rob} reports, for the three models on which the full attack suite was run (Qwen3.5-4B has injection data only), the post-attack FSR (\textbf{P}) and the retention (\textbf{R}~= post/pre) for every attack, so that both the absolute surviving capability and the relative retention are visible. Within each model block, the six methods' rows are grouped per language so that methods can be compared directly under the same condition. Table~\ref{tab:app-ppl-perlang} gives the per-language LCF template perplexity underlying the aggregate in Fig.~\ref{fig:ppl}.

\begingroup
\small
\setlength{\tabcolsep}{5pt}
\begin{longtable}{llcccccc}
\caption{Per-language injection FSR (\%) for all four models and six injection methods. LCFEdit and AlphaEdit are our measurements; MCEdit, FPEdit, PREE, and LoRA are per-language figures from the same comparison suite.}\label{tab:app-inj}\\
\toprule
Code & Language & LCFEdit & AlphaEdit & MCEdit & FPEdit & PREE & LoRA \\
\midrule
\endfirsthead
\multicolumn{8}{c}{\tablename\ \thetable\ -- continued}\\
\toprule
Code & Language & LCFEdit & AlphaEdit & MCEdit & FPEdit & PREE & LoRA \\
\midrule
\endhead
\midrule\multicolumn{8}{r}{\textit{continued on next page}}\\
\endfoot
\bottomrule
\endlastfoot
\multicolumn{8}{l}{\textbf{Qwen3.5-9B}}\\
\midrule
bn & Bengali & 100.00 & 63.64 & 100.00 & 100.00 & 81.82 & 54.55 \\
fa & Persian & 100.00 & 90.91 & 100.00 & 90.91 & 90.91 & 45.45 \\
hi & Hindi & 97.30 & 36.36 & 90.91 & 90.91 & 90.91 & 54.55 \\
id & Indonesian & 100.00 & 81.82 & 100.00 & 90.91 & 90.91 & 54.55 \\
it & Italian & 100.00 & 81.82 & 100.00 & 90.91 & 90.91 & 45.45 \\
nl & Dutch & 100.00 & 45.45 & 100.00 & 100.00 & 90.91 & 45.45 \\
pl & Polish & 100.00 & 90.91 & 100.00 & 90.91 & 90.91 & 45.45 \\
pt & Portuguese & 100.00 & 81.82 & 100.00 & 90.91 & 90.91 & 45.45 \\
sv & Swedish & 100.00 & 54.55 & 100.00 & 100.00 & 90.91 & 54.55 \\
th & Thai & 97.30 & 45.45 & 90.91 & 90.91 & 90.91 & 45.45 \\
tr & Turkish & 100.00 & 72.73 & 100.00 & 100.00 & 90.91 & 54.55 \\
vi & Vietnamese & 100.00 & 90.91 & 100.00 & 100.00 & 90.91 & 45.45 \\
\textbf{Mean} &  & \textbf{99.55} & \textbf{69.70} & \textbf{98.48} & \textbf{94.70} & \textbf{90.15} & \textbf{49.24} \\
\midrule
\multicolumn{8}{l}{\textbf{Qwen3.5-4B}}\\
\midrule
bn & Bengali & 100.00 & 63.64 & 100.00 & 100.00 & 90.91 & 45.45 \\
fa & Persian & 100.00 & 72.73 & 100.00 & 100.00 & 90.91 & 45.45 \\
hi & Hindi & 90.91 & 24.90 & 81.82 & 81.82 & 81.82 & 45.45 \\
id & Indonesian & 100.00 & 72.73 & 100.00 & 100.00 & 90.91 & 54.55 \\
it & Italian & 100.00 & 63.64 & 100.00 & 90.91 & 90.91 & 45.45 \\
nl & Dutch & 100.00 & 45.45 & 100.00 & 100.00 & 90.91 & 54.55 \\
pl & Polish & 100.00 & 90.91 & 100.00 & 100.00 & 100.00 & 45.45 \\
pt & Portuguese & 90.91 & 81.82 & 81.82 & 81.82 & 81.82 & 54.55 \\
sv & Swedish & 100.00 & 36.30 & 100.00 & 90.91 & 90.91 & 54.55 \\
th & Thai & 100.00 & 45.45 & 100.00 & 90.91 & 90.91 & 54.55 \\
tr & Turkish & 100.00 & 72.73 & 100.00 & 100.00 & 90.91 & 45.45 \\
vi & Vietnamese & 100.00 & 81.82 & 100.00 & 100.00 & 90.91 & 45.45 \\
\textbf{Mean} &  & \textbf{98.48} & \textbf{62.68} & \textbf{96.97} & \textbf{94.70} & \textbf{90.15} & \textbf{49.24} \\
\midrule
\multicolumn{8}{l}{\textbf{Qwen3.5-0.8B}}\\
\midrule
bn & Bengali & 83.00 & 52.80 & 81.82 & 81.82 & 81.82 & 45.45 \\
fa & Persian & 83.00 & 57.20 & 81.82 & 81.82 & 81.82 & 45.45 \\
hi & Hindi & 80.00 & 36.36 & 81.82 & 72.73 & 72.73 & 45.45 \\
id & Indonesian & 95.00 & 68.60 & 90.91 & 90.91 & 90.91 & 54.55 \\
it & Italian & 100.00 & 45.40 & 100.00 & 100.00 & 90.91 & 54.55 \\
nl & Dutch & 100.00 & 45.45 & 100.00 & 100.00 & 90.91 & 54.55 \\
pl & Polish & 100.00 & 81.82 & 100.00 & 100.00 & 90.91 & 45.45 \\
pt & Portuguese & 91.00 & 74.40 & 90.91 & 90.91 & 90.91 & 45.45 \\
sv & Swedish & 88.00 & 38.90 & 81.82 & 81.82 & 81.82 & 45.45 \\
th & Thai & 100.00 & 36.36 & 100.00 & 100.00 & 81.82 & 54.55 \\
tr & Turkish & 100.00 & 54.55 & 100.00 & 100.00 & 90.91 & 45.45 \\
vi & Vietnamese & 100.00 & 90.91 & 100.00 & 100.00 & 81.82 & 54.55 \\
\textbf{Mean} &  & \textbf{93.33} & \textbf{56.90} & \textbf{92.42} & \textbf{91.67} & \textbf{85.61} & \textbf{49.24} \\
\midrule
\multicolumn{8}{l}{\textbf{Llama-3.2-3B-Instruct}}\\
\midrule
bn & Bengali & 63.64 & 27.27 & 63.64 & 54.55 & 54.55 & 54.55 \\
fa & Persian & 100.00 & 72.73 & 100.00 & 90.91 & 90.91 & 45.45 \\
hi & Hindi & 81.82 & 9.09 & 72.73 & 72.73 & 72.73 & 45.45 \\
id & Indonesian & 90.91 & 54.55 & 81.82 & 81.82 & 81.82 & 45.45 \\
it & Italian & 100.00 & 72.73 & 100.00 & 100.00 & 100.00 & 45.45 \\
nl & Dutch & 94.40 & 36.36 & 90.91 & 90.91 & 90.91 & 54.55 \\
pl & Polish & 100.00 & 81.82 & 100.00 & 100.00 & 90.91 & 54.55 \\
pt & Portuguese & 85.00 & 81.82 & 81.82 & 81.82 & 81.82 & 54.55 \\
sv & Swedish & 100.00 & 36.36 & 100.00 & 100.00 & 100.00 & 54.55 \\
th & Thai & 100.00 & 45.45 & 100.00 & 100.00 & 100.00 & 45.45 \\
tr & Turkish & 100.00 & 63.64 & 100.00 & 100.00 & 100.00 & 45.45 \\
vi & Vietnamese & 100.00 & 81.82 & 90.91 & 100.00 & 100.00 & 45.45 \\
\textbf{Mean} &  & \textbf{92.98} & \textbf{55.30} & \textbf{90.66} & \textbf{89.14} & \textbf{88.64} & \textbf{49.24} \\
\end{longtable}
\endgroup

\begingroup
\scriptsize
\setlength{\tabcolsep}{2.4pt}
\begin{longtable}{llccccccccccccc}
\caption{Per-language detectability under six attacks. Each attack reports post-attack FSR (\textbf{P}, \%) and retention (\textbf{R} = post/pre, \%). ``Pre'' is the injection FSR before attack. Pr30/Pr40: 30/40\% L1 pruning; FT-A/FT-M: LoRA fine-tuning on Alpaca / MathInstruct. Blocks are grouped by model; within each model, the six methods' rows (LCFEdit ours; AlphaEdit, MCEdit, FPEdit, PREE, and LoRA baselines) are interleaved per language for direct same-condition comparison. Only the per-model mean rows are set in bold, for at-a-glance comparison.}\label{tab:app-rob}\\
\toprule
Method & Lang & Pre & \multicolumn{2}{c}{INT8} & \multicolumn{2}{c}{NF4} & \multicolumn{2}{c}{Pr30} & \multicolumn{2}{c}{Pr40} & \multicolumn{2}{c}{FT-A} & \multicolumn{2}{c}{FT-M} \\
 & & FSR & P & R & P & R & P & R & P & R & P & R & P & R \\
\cmidrule(lr){4-5}\cmidrule(lr){6-7}\cmidrule(lr){8-9}\cmidrule(lr){10-11}\cmidrule(lr){12-13}\cmidrule(lr){14-15}
\endfirsthead
\multicolumn{15}{c}{\tablename\ \thetable\ -- continued}\\
\toprule
Method & Lang & Pre & \multicolumn{2}{c}{INT8} & \multicolumn{2}{c}{NF4} & \multicolumn{2}{c}{Pr30} & \multicolumn{2}{c}{Pr40} & \multicolumn{2}{c}{FT-A} & \multicolumn{2}{c}{FT-M} \\
 & & FSR & P & R & P & R & P & R & P & R & P & R & P & R \\
\cmidrule(lr){4-5}\cmidrule(lr){6-7}\cmidrule(lr){8-9}\cmidrule(lr){10-11}\cmidrule(lr){12-13}\cmidrule(lr){14-15}
\endhead
\midrule\multicolumn{15}{r}{\textit{continued on next page}}\\
\endfoot
\bottomrule
\endlastfoot
\multicolumn{15}{l}{\textbf{Qwen3.5-9B}}\\
\midrule
LCFEdit   & bn & 100.00 & 100.00 & 100.00 & 100.00 & 100.00 & 79.30 & 79.30 & 9.09  & 9.09  & 90.91 & 90.91 & 97.10 & 97.10           \\
AlphaEdit & bn & 63.64  & 63.64           & 100.00 & 63.64           & 100.00 & 33.90          & 53.30          & 0.00           & 0.00           & 48.90          & 76.90          & 63.64          & 100.00 \\
MCEdit & bn & 100.00 & 100.00 & 100.00 & 100.00 & 100.00 & 72.73 & 72.73 & 9.09 & 9.09 & 81.82 & 81.82 & 90.91 & 90.91 \\
FPEdit & bn & 100.00 & 90.91 & 90.91 & 90.91 & 90.91 & 72.73 & 72.73 & 9.09 & 9.09 & 81.82 & 81.82 & 90.91 & 90.91 \\
PREE & bn & 81.82 & 81.82 & 100.00 & 72.73 & 88.89 & 54.55 & 66.67 & 9.09 & 11.11 & 72.73 & 88.89 & 81.82 & 100.00 \\
LoRA & bn & 54.55 & 72.73 & 133.33 & 45.45 & 83.32 & 36.36 & 66.65 & 9.09 & 16.66 & 54.55 & 100.00 & 54.55 & 100.00 \\
LCFEdit   & fa & 100.00 & 62.80           & 62.80           & 62.80           & 62.80           & 59.50 & 59.50 & 15.70 & 15.70 & 43.60          & 43.60          & 54.70          & 54.70           \\
AlphaEdit & fa & 90.91  & 75.20  & 82.70  & 71.10  & 78.20  & 49.60          & 54.60          & 10.70          & 11.80          & 61.10 & 67.20 & 63.30 & 69.60  \\
MCEdit & fa & 100.00 & 54.55 & 54.55 & 54.55 & 54.55 & 54.55 & 54.55 & 9.09 & 9.09 & 36.36 & 36.36 & 54.55 & 54.55 \\
FPEdit & fa & 90.91 & 54.55 & 60.00 & 54.55 & 60.00 & 54.55 & 60.00 & 9.09 & 10.00 & 36.36 & 40.00 & 54.55 & 60.00 \\
PREE & fa & 90.91 & 54.55 & 60.00 & 54.55 & 60.00 & 54.55 & 60.00 & 9.09 & 10.00 & 36.36 & 40.00 & 54.55 & 60.00 \\
LoRA & fa & 45.45 & 54.55 & 120.02 & 54.55 & 120.02 & 36.36 & 80.00 & 9.09 & 20.00 & 36.36 & 80.00 & 54.55 & 120.02 \\
LCFEdit   & hi & 97.30  & 90.91  & 93.40           & 90.91  & 93.40           & 24.00 & 24.70 & 0.00  & 0.00  & 68.70 & 70.60 & 85.80 & 88.20  \\
AlphaEdit & hi & 36.36  & 36.36           & 100.00 & 36.36           & 100.00 & 0.00           & 0.00           & 0.00  & 0.00  & 21.30          & 58.50          & 6.70           & 18.40           \\
MCEdit & hi & 90.91 & 81.82 & 90.00 & 81.82 & 90.00 & 18.18 & 20.00 & 0.00 & 0.00 & 63.64 & 70.00 & 81.82 & 90.00 \\
FPEdit & hi & 90.91 & 81.82 & 90.00 & 81.82 & 90.00 & 18.18 & 20.00 & 0.00 & 0.00 & 63.64 & 70.00 & 81.82 & 90.00 \\
PREE & hi & 90.91 & 81.82 & 90.00 & 72.73 & 80.00 & 18.18 & 20.00 & 0.00 & 0.00 & 63.64 & 70.00 & 72.73 & 80.00 \\
LoRA & hi & 54.55 & 72.73 & 133.33 & 54.55 & 100.00 & 18.18 & 33.33 & 0.00 & 0.00 & 54.55 & 100.00 & 54.55 & 100.00 \\
LCFEdit   & id & 100.00 & 100.00 & 100.00 & 100.00 & 100.00 & 93.40 & 93.40 & 58.70 & 58.70 & 99.80 & 99.80 & 100.00& 100.00 \\
AlphaEdit & id & 81.82  & 78.50           & 96.00           & 80.20           & 98.00           & 64.50          & 78.90          & 40.50          & 49.50          & 72.90          & 89.10          & 81.30          & 99.40           \\
MCEdit & id & 100.00 & 100.00 & 100.00 & 100.00 & 100.00 & 90.91 & 90.91 & 36.36 & 36.36 & 90.91 & 90.91 & 100.00 & 100.00 \\
FPEdit & id & 90.91 & 100.00 & 110.00 & 100.00 & 110.00 & 72.73 & 80.00 & 18.18 & 20.00 & 90.91 & 100.00 & 90.91 & 100.00 \\
PREE & id & 90.91 & 81.82 & 90.00 & 72.73 & 80.00 & 54.55 & 60.00 & 9.09 & 10.00 & 72.73 & 80.00 & 81.82 & 90.00 \\
LoRA & id & 54.55 & 72.73 & 133.33 & 54.55 & 100.00 & 36.36 & 66.65 & 9.09 & 16.66 & 45.45 & 83.32 & 63.64 & 116.66 \\
LCFEdit   & it & 100.00 & 100.00 & 100.00 & 100.00 & 100.00 & 86.80 & 86.80 & 14.90          & 14.90          & 100.00& 100.00& 100.00& 100.00 \\
AlphaEdit & it & 81.82  & 79.30           & 96.90           & 76.90           & 94.00           & 61.20          & 74.80          & 17.40 & 21.30 & 74.90          & 91.60          & 72.73          & 88.90           \\
MCEdit & it & 100.00 & 100.00 & 100.00 & 100.00 & 100.00 & 81.82 & 81.82 & 9.09 & 9.09 & 100.00 & 100.00 & 100.00 & 100.00 \\
FPEdit & it & 90.91 & 100.00 & 110.00 & 90.91 & 100.00 & 72.73 & 80.00 & 9.09 & 10.00 & 90.91 & 100.00 & 81.82 & 90.00 \\
PREE & it & 90.91 & 72.73 & 80.00 & 72.73 & 80.00 & 54.55 & 60.00 & 9.09 & 10.00 & 72.73 & 80.00 & 72.73 & 80.00 \\
LoRA & it & 45.45 & 63.64 & 140.02 & 54.55 & 120.02 & 45.45 & 100.00 & 9.09 & 20.00 & 54.55 & 120.02 & 63.64 & 140.02 \\
LCFEdit   & nl & 100.00 & 100.00 & 100.00 & 100.00 & 100.00 & 72.73 & 72.73 & 9.09  & 9.09  & 100.00& 100.00& 100.00& 100.00 \\
AlphaEdit & nl & 45.45  & 45.45           & 100.00 & 45.45           & 100.00 & 27.27          & 60.00          & 0.00           & 0.00           & 45.45          & 100.00& 44.00          & 96.70           \\
MCEdit & nl & 100.00 & 100.00 & 100.00 & 100.00 & 100.00 & 63.64 & 63.64 & 9.09 & 9.09 & 100.00 & 100.00 & 100.00 & 100.00 \\
FPEdit & nl & 100.00 & 100.00 & 100.00 & 90.91 & 90.91 & 63.64 & 63.64 & 9.09 & 9.09 & 90.91 & 90.91 & 90.91 & 90.91 \\
PREE & nl & 90.91 & 81.82 & 90.00 & 72.73 & 80.00 & 54.55 & 60.00 & 9.09 & 10.00 & 63.64 & 70.00 & 72.73 & 80.00 \\
LoRA & nl & 45.45 & 72.73 & 160.02 & 54.55 & 120.02 & 36.36 & 80.00 & 9.09 & 20.00 & 54.55 & 120.02 & 63.64 & 140.02 \\
LCFEdit   & pl & 100.00 & 100.00 & 100.00 & 100.00 & 100.00 & 76.90 & 76.90 & 9.09  & 9.09  & 99.80 & 99.80 & 100.00& 100.00 \\
AlphaEdit & pl & 90.91  & 74.40           & 81.80           & 77.70           & 85.50           & 20.70          & 22.80          & 3.30           & 3.60           & 48.20          & 53.00          & 76.90          & 84.60           \\
MCEdit & pl & 100.00 & 100.00 & 100.00 & 100.00 & 100.00 & 72.73 & 72.73 & 9.09 & 9.09 & 90.91 & 90.91 & 100.00 & 100.00 \\
FPEdit & pl & 90.91 & 100.00 & 110.00 & 90.91 & 100.00 & 72.73 & 80.00 & 9.09 & 10.00 & 90.91 & 100.00 & 90.91 & 100.00 \\
PREE & pl & 90.91 & 81.82 & 90.00 & 72.73 & 80.00 & 54.55 & 60.00 & 9.09 & 10.00 & 72.73 & 80.00 & 72.73 & 80.00 \\
LoRA & pl & 45.45 & 63.64 & 140.02 & 54.55 & 120.02 & 45.45 & 100.00 & 9.09 & 20.00 & 54.55 & 120.02 & 63.64 & 140.02 \\
LCFEdit   & pt & 100.00 & 90.91  & 90.91  & 90.91  & 90.91  & 55.40 & 55.40 & 1.70  & 1.70           & 60.00          & 60.00          & 90.91 & 90.91  \\
AlphaEdit & pt & 81.82  & 67.80           & 82.90           & 67.80           & 82.90           & 38.80          & 47.40          & 1.70  & 2.10  & 61.10 & 74.70 & 61.30          & 74.90           \\
MCEdit & pt & 100.00 & 81.82 & 81.82 & 81.82 & 81.82 & 54.55 & 54.55 & 0.00 & 0.00 & 54.55 & 54.55 & 81.82 & 81.82 \\
FPEdit & pt & 90.91 & 81.82 & 90.00 & 81.82 & 90.00 & 54.55 & 60.00 & 0.00 & 0.00 & 54.55 & 60.00 & 81.82 & 90.00 \\
PREE & pt & 90.91 & 81.82 & 90.00 & 72.73 & 80.00 & 54.55 & 60.00 & 0.00 & 0.00 & 54.55 & 60.00 & 72.73 & 80.00 \\
LoRA & pt & 45.45 & 63.64 & 140.02 & 45.45 & 100.00 & 45.45 & 100.00 & 0.00 & 0.00 & 54.55 & 120.02 & 54.55 & 120.02 \\
LCFEdit   & sv & 100.00 & 100.00 & 100.00 & 95.00  & 95.00           & 72.73 & 72.73 & 1.70  & 1.70  & 100.00& 100.00& 100.00& 100.00 \\
AlphaEdit & sv & 54.55  & 53.70           & 98.50           & 54.55           & 100.00 & 24.80          & 45.50          & 0.80           & 1.50           & 42.90          & 78.70          & 53.80          & 98.70           \\
MCEdit & sv & 100.00 & 100.00 & 100.00 & 90.91 & 90.91 & 63.64 & 63.64 & 0.00 & 0.00 & 100.00 & 100.00 & 100.00 & 100.00 \\
FPEdit & sv & 100.00 & 90.91 & 90.91 & 90.91 & 90.91 & 63.64 & 63.64 & 0.00 & 0.00 & 90.91 & 90.91 & 90.91 & 90.91 \\
PREE & sv & 90.91 & 81.82 & 90.00 & 72.73 & 80.00 & 54.55 & 60.00 & 0.00 & 0.00 & 72.73 & 80.00 & 72.73 & 80.00 \\
LoRA & sv & 54.55 & 63.64 & 116.66 & 54.55 & 100.00 & 36.36 & 66.65 & 0.00 & 0.00 & 54.55 & 100.00 & 63.64 & 116.66 \\
LCFEdit   & th & 97.30  & 97.30  & 100.00 & 97.30  & 100.00 & 19.00 & 19.50 & 0.00  & 0.00  & 92.20 & 94.80          & 97.30 & 100.00 \\
AlphaEdit & th & 45.45  & 44.60           & 98.00           & 39.70           & 87.30           & 8.30           & 18.20          & 0.00  & 0.00  & 44.50          & 97.80 & 45.45          & 100.00 \\
MCEdit & th & 90.91 & 90.91 & 100.00 & 90.91 & 100.00 & 18.18 & 20.00 & 0.00 & 0.00 & 90.91 & 100.00 & 90.91 & 100.00 \\
FPEdit & th & 90.91 & 90.91 & 100.00 & 90.91 & 100.00 & 18.18 & 20.00 & 0.00 & 0.00 & 81.82 & 90.00 & 81.82 & 90.00 \\
PREE & th & 90.91 & 81.82 & 90.00 & 72.73 & 80.00 & 18.18 & 20.00 & 0.00 & 0.00 & 72.73 & 80.00 & 81.82 & 90.00 \\
LoRA & th & 45.45 & 63.64 & 140.02 & 54.55 & 120.02 & 18.18 & 40.00 & 0.00 & 0.00 & 54.55 & 120.02 & 63.64 & 140.02 \\
LCFEdit   & tr & 100.00 & 100.00 & 100.00 & 100.00 & 100.00 & 86.80 & 86.80 & 39.70 & 39.70 & 99.50 & 99.50 & 100.00& 100.00 \\
AlphaEdit & tr & 72.73  & 69.40           & 95.50           & 60.30           & 82.90           & 25.60          & 35.20          & 6.60           & 9.10           & 48.20          & 66.30          & 58.00          & 79.80           \\
MCEdit & tr & 100.00 & 100.00 & 100.00 & 100.00 & 100.00 & 81.82 & 81.82 & 27.27 & 27.27 & 90.91 & 90.91 & 100.00 & 100.00 \\
FPEdit & tr & 100.00 & 90.91 & 90.91 & 90.91 & 90.91 & 72.73 & 72.73 & 18.18 & 18.18 & 81.82 & 81.82 & 90.91 & 90.91 \\
PREE & tr & 90.91 & 72.73 & 80.00 & 72.73 & 80.00 & 54.55 & 60.00 & 9.09 & 10.00 & 72.73 & 80.00 & 81.82 & 90.00 \\
LoRA & tr & 54.55 & 72.73 & 133.33 & 54.55 & 100.00 & 45.45 & 83.32 & 9.09 & 16.66 & 54.55 & 100.00 & 63.64 & 116.66 \\
LCFEdit   & vi & 100.00 & 100.00 & 100.00 & 100.00 & 100.00 & 90.91 & 90.91 & 43.80 & 43.80 & 98.70 & 98.70 & 100.00& 100.00 \\
AlphaEdit & vi & 90.91  & 89.30           & 98.20           & 86.00           & 94.60           & 69.40          & 76.30          & 11.60          & 12.80          & 66.20          & 72.80          & 80.70          & 88.80           \\
MCEdit & vi & 100.00 & 100.00 & 100.00 & 100.00 & 100.00 & 81.82 & 81.82 & 27.27 & 27.27 & 90.91 & 90.91 & 100.00 & 100.00 \\
FPEdit & vi & 100.00 & 100.00 & 100.00 & 100.00 & 100.00 & 81.82 & 81.82 & 18.18 & 18.18 & 81.82 & 81.82 & 90.91 & 90.91 \\
PREE & vi & 90.91 & 81.82 & 90.00 & 72.73 & 80.00 & 54.55 & 60.00 & 18.18 & 20.00 & 72.73 & 80.00 & 72.73 & 80.00 \\
LoRA & vi & 45.45 & 63.64 & 140.02 & 54.55 & 120.02 & 45.45 & 100.00 & 18.18 & 40.00 & 54.55 & 120.02 & 54.55 & 120.02 \\
\textbf{LCFEdit}   & \textbf{Mean} & \textbf{99.55} & \textbf{95.16} & \textbf{95.59} & \textbf{94.74} & \textbf{95.18} & \textbf{68.12} & \textbf{68.22} & \textbf{16.96} & \textbf{16.96} & \textbf{87.77} & \textbf{88.14} & \textbf{93.82} & \textbf{94.24} \\
\textbf{AlphaEdit} & \textbf{Mean} & \textbf{69.70} & 64.80          & 94.21          & 63.31          & 91.95          & 35.34          & 47.25          & 7.72           & 9.31           & 52.97          & 77.22          & 58.99          & 83.32          \\
\textbf{MCEdit} & \textbf{Mean} & \textbf{98.48} & 92.42 & 93.85 & 91.67 & 93.08 & 62.88 & 63.85 & 11.36 & 11.54 & 82.58 & 83.85 & 91.67 & 93.08 \\
\textbf{FPEdit} & \textbf{Mean} & \textbf{94.70} & 90.15 & 95.20 & 87.88 & 92.80 & 59.85 & 63.20 & 8.33 & 8.80 & 78.03 & 82.40 & 84.85 & 89.60 \\
\textbf{PREE} & \textbf{Mean} & \textbf{90.15} & 78.03 & 86.56 & 71.21 & 78.99 & 48.48 & 53.78 & 6.82 & 7.57 & 66.67 & 73.95 & 74.24 & 82.35 \\
\textbf{LoRA} & \textbf{Mean} & \textbf{49.24} & 66.67 & 135.40 & 53.03 & 107.70 & 37.12 & 75.39 & 6.82 & 13.85 & 52.27 & 106.15 & 59.85 & 121.55 \\
\midrule
\multicolumn{15}{l}{\textbf{Qwen3.5-0.8B}}\\
\midrule
LCFEdit   & bn & 83.00  & 69.10  & 83.30  & 62.10  & 74.80  & 52.20 & 62.90 & 5.00  & 6.00  & 59.30 & 71.40 & 59.20 & 71.30  \\
AlphaEdit & bn & 52.80  & 36.20           & 68.60           & 25.20           & 47.70           & 22.70          & 43.00          & 1.00           & 1.90           & 23.70          & 44.90          & 35.50          & 67.20           \\
MCEdit & bn & 81.82 & 72.73 & 88.89 & 63.64 & 77.78 & 54.55 & 66.67 & 9.09 & 11.11 & 63.64 & 77.78 & 63.64 & 77.78 \\
FPEdit & bn & 81.82 & 63.64 & 77.78 & 54.55 & 66.67 & 45.45 & 55.55 & 0.00 & 0.00 & 54.55 & 66.67 & 54.55 & 66.67 \\
PREE & bn & 81.82 & 63.64 & 77.78 & 54.55 & 66.67 & 45.45 & 55.55 & 0.00 & 0.00 & 54.55 & 66.67 & 54.55 & 66.67 \\
LoRA & bn & 45.45 & 63.64 & 140.02 & 54.55 & 120.02 & 36.36 & 80.00 & 0.00 & 0.00 & 54.55 & 120.02 & 54.55 & 120.02 \\
LCFEdit   & fa & 83.00  & 50.30  & 60.60  & 33.90  & 40.80           & 43.00 & 51.80 & 7.30  & 8.80  & 18.18          & 21.90          & 34.60 & 41.70           \\
AlphaEdit & fa & 57.20  & 31.50           & 55.10           & 30.80           & 53.80  & 18.90          & 33.00          & 2.60           & 4.50           & 23.40 & 40.90 & 24.10          & 42.10  \\
MCEdit & fa & 81.82 & 45.45 & 55.55 & 27.27 & 33.33 & 36.36 & 44.44 & 0.00 & 0.00 & 18.18 & 22.22 & 27.27 & 33.33 \\
FPEdit & fa & 81.82 & 45.45 & 55.55 & 27.27 & 33.33 & 36.36 & 44.44 & 0.00 & 0.00 & 18.18 & 22.22 & 27.27 & 33.33 \\
PREE & fa & 81.82 & 45.45 & 55.55 & 27.27 & 33.33 & 36.36 & 44.44 & 0.00 & 0.00 & 18.18 & 22.22 & 27.27 & 33.33 \\
LoRA & fa & 45.45 & 45.45 & 100.00 & 27.27 & 60.00 & 36.36 & 80.00 & 0.00 & 0.00 & 18.18 & 40.00 & 27.27 & 60.00 \\
LCFEdit   & hi & 80.00  & 66.90  & 83.60  & 60.10  & 75.10  & 13.60 & 17.00 & 0.00  & 0.00  & 41.40 & 51.70 & 44.60 & 55.80  \\
AlphaEdit & hi & 36.36  & 27.60           & 75.80           & 21.70           & 59.60           & 2.60           & 7.10           & 0.00  & 0.00  & 10.00          & 27.50          & 3.50           & 9.60            \\
MCEdit & hi & 81.82 & 63.64 & 77.78 & 54.55 & 66.67 & 9.09 & 11.11 & 0.00 & 0.00 & 36.36 & 44.44 & 36.36 & 44.44 \\
FPEdit & hi & 72.73 & 63.64 & 87.50 & 54.55 & 75.00 & 9.09 & 12.50 & 0.00 & 0.00 & 36.36 & 49.99 & 36.36 & 49.99 \\
PREE & hi & 72.73 & 63.64 & 87.50 & 54.55 & 75.00 & 9.09 & 12.50 & 0.00 & 0.00 & 36.36 & 49.99 & 36.36 & 49.99 \\
LoRA & hi & 45.45 & 63.64 & 140.02 & 54.55 & 120.02 & 9.09 & 20.00 & 0.00 & 0.00 & 36.36 & 80.00 & 36.36 & 80.00 \\
LCFEdit   & id & 95.00  & 79.50  & 83.70  & 64.00  & 67.40  & 68.90 & 72.50 & 29.00 & 30.50 & 57.30 & 60.30 & 85.00 & 89.50  \\
AlphaEdit & id & 68.60  & 37.00           & 53.90           & 36.60           & 53.40           & 24.10          & 35.10          & 10.20          & 14.90          & 31.40          & 45.80          & 29.60          & 43.10           \\
MCEdit & id & 90.91 & 72.73 & 80.00 & 63.64 & 70.00 & 63.64 & 70.00 & 27.27 & 30.00 & 54.55 & 60.00 & 81.82 & 90.00 \\
FPEdit & id & 90.91 & 72.73 & 80.00 & 63.64 & 70.00 & 63.64 & 70.00 & 27.27 & 30.00 & 54.55 & 60.00 & 81.82 & 90.00 \\
PREE & id & 90.91 & 72.73 & 80.00 & 63.64 & 70.00 & 54.55 & 60.00 & 27.27 & 30.00 & 54.55 & 60.00 & 81.82 & 90.00 \\
LoRA & id & 54.55 & 72.73 & 133.33 & 54.55 & 100.00 & 45.45 & 83.32 & 27.27 & 49.99 & 54.55 & 100.00 & 72.73 & 133.33 \\
LCFEdit   & it & 100.00 & 96.40  & 96.40  & 74.60  & 74.60  & 66.90 & 66.90 & 8.90  & 8.90  & 68.70 & 68.70 & 90.20 & 90.20  \\
AlphaEdit & it & 45.40  & 32.60           & 71.80           & 17.50           & 38.50           & 20.70          & 45.60          & 1.90           & 4.20           & 25.80          & 56.80          & 15.20          & 33.50           \\
MCEdit & it & 100.00 & 90.91 & 90.91 & 72.73 & 72.73 & 63.64 & 63.64 & 0.00 & 0.00 & 63.64 & 63.64 & 81.82 & 81.82 \\
FPEdit & it & 100.00 & 90.91 & 90.91 & 72.73 & 72.73 & 63.64 & 63.64 & 0.00 & 0.00 & 63.64 & 63.64 & 81.82 & 81.82 \\
PREE & it & 90.91 & 72.73 & 80.00 & 72.73 & 80.00 & 45.45 & 49.99 & 0.00 & 0.00 & 63.64 & 70.00 & 81.82 & 90.00 \\
LoRA & it & 54.55 & 63.64 & 116.66 & 54.55 & 100.00 & 45.45 & 83.32 & 0.00 & 0.00 & 63.64 & 116.66 & 72.73 & 133.33 \\
LCFEdit   & nl & 100.00 & 97.30  & 97.30  & 69.00  & 69.00  & 64.40 & 64.40 & 4.60  & 4.60  & 65.50 & 65.50 & 84.20 & 84.20  \\
AlphaEdit & nl & 45.45  & 25.30           & 55.60           & 20.40           & 44.80           & 18.40          & 40.40          & 1.20           & 2.60           & 29.50          & 64.80          & 24.00          & 52.70           \\
MCEdit & nl & 100.00 & 90.91 & 90.91 & 63.64 & 63.64 & 63.64 & 63.64 & 0.00 & 0.00 & 63.64 & 63.64 & 81.82 & 81.82 \\
FPEdit & nl & 100.00 & 90.91 & 90.91 & 63.64 & 63.64 & 63.64 & 63.64 & 0.00 & 0.00 & 63.64 & 63.64 & 81.82 & 81.82 \\
PREE & nl & 90.91 & 72.73 & 80.00 & 63.64 & 70.00 & 45.45 & 49.99 & 0.00 & 0.00 & 63.64 & 70.00 & 81.82 & 90.00 \\
LoRA & nl & 54.55 & 72.73 & 133.33 & 54.55 & 100.00 & 45.45 & 83.32 & 0.00 & 0.00 & 63.64 & 116.66 & 81.82 & 149.99 \\
LCFEdit   & pl & 100.00 & 87.50  & 87.50  & 87.00  & 87.00  & 63.30 & 63.30 & 4.40  & 4.40  & 54.00 & 54.00 & 65.80 & 65.80  \\
AlphaEdit & pl & 81.82  & 37.50           & 45.80           & 29.00           & 35.50           & 21.40          & 26.20          & 1.40           & 1.70           & 17.80          & 21.80          & 27.40          & 33.50           \\
MCEdit & pl & 100.00 & 81.82 & 81.82 & 81.82 & 81.82 & 54.55 & 54.55 & 0.00 & 0.00 & 45.45 & 45.45 & 63.64 & 63.64 \\
FPEdit & pl & 100.00 & 81.82 & 81.82 & 81.82 & 81.82 & 54.55 & 54.55 & 0.00 & 0.00 & 45.45 & 45.45 & 63.64 & 63.64 \\
PREE & pl & 90.91 & 81.82 & 90.00 & 72.73 & 80.00 & 45.45 & 49.99 & 0.00 & 0.00 & 45.45 & 49.99 & 63.64 & 70.00 \\
LoRA & pl & 45.45 & 72.73 & 160.02 & 54.55 & 120.02 & 45.45 & 100.00 & 0.00 & 0.00 & 45.45 & 100.00 & 63.64 & 140.02 \\
LCFEdit   & pt & 91.00  & 79.70  & 87.60  & 75.00  & 82.40  & 40.60 & 44.60 & 1.30  & 1.40  & 38.80 & 42.60 & 50.90 & 55.90  \\
AlphaEdit & pt & 74.40  & 42.90           & 57.70           & 41.20           & 55.40           & 19.00          & 25.50          & 0.40           & 0.50           & 30.10          & 40.50          & 30.40          & 40.90           \\
MCEdit & pt & 90.91 & 72.73 & 80.00 & 72.73 & 80.00 & 36.36 & 40.00 & 0.00 & 0.00 & 36.36 & 40.00 & 45.45 & 49.99 \\
FPEdit & pt & 90.91 & 72.73 & 80.00 & 72.73 & 80.00 & 36.36 & 40.00 & 0.00 & 0.00 & 36.36 & 40.00 & 45.45 & 49.99 \\
PREE & pt & 90.91 & 72.73 & 80.00 & 72.73 & 80.00 & 36.36 & 40.00 & 0.00 & 0.00 & 36.36 & 40.00 & 45.45 & 49.99 \\
LoRA & pt & 45.45 & 72.73 & 160.02 & 54.55 & 120.02 & 36.36 & 80.00 & 0.00 & 0.00 & 36.36 & 80.00 & 45.45 & 100.00 \\
LCFEdit   & sv & 88.00  & 73.80  & 83.90  & 60.00  & 68.20  & 54.90 & 62.40 & 1.00  & 1.10  & 54.90 & 62.40 & 73.80 & 83.90  \\
AlphaEdit & sv & 38.90  & 21.30           & 54.80           & 18.18           & 46.80           & 12.60          & 32.40          & 0.10           & 0.30           & 17.90          & 46.00          & 18.18          & 46.80           \\
MCEdit & sv & 81.82 & 72.73 & 88.89 & 54.55 & 66.67 & 54.55 & 66.67 & 0.00 & 0.00 & 54.55 & 66.67 & 72.73 & 88.89 \\
FPEdit & sv & 81.82 & 72.73 & 88.89 & 54.55 & 66.67 & 54.55 & 66.67 & 0.00 & 0.00 & 54.55 & 66.67 & 72.73 & 88.89 \\
PREE & sv & 81.82 & 72.73 & 88.89 & 54.55 & 66.67 & 45.45 & 55.55 & 0.00 & 0.00 & 54.55 & 66.67 & 72.73 & 88.89 \\
LoRA & sv & 45.45 & 72.73 & 160.02 & 54.55 & 120.02 & 45.45 & 100.00 & 0.00 & 0.00 & 54.55 & 120.02 & 72.73 & 160.02 \\
LCFEdit   & th & 100.00 & 99.10  & 99.10  & 84.40  & 84.40  & 16.30 & 16.30 & 0.00  & 0.00  & 69.30 & 69.30 & 67.00 & 67.00  \\
AlphaEdit & th & 36.36  & 26.60           & 73.10           & 14.80           & 40.70           & 3.90           & 10.70          & 0.00  & 0.00  & 20.90          & 57.40          & 18.10          & 49.70           \\
MCEdit & th & 100.00 & 90.91 & 90.91 & 81.82 & 81.82 & 9.09 & 9.09 & 0.00 & 0.00 & 63.64 & 63.64 & 63.64 & 63.64 \\
FPEdit & th & 100.00 & 90.91 & 90.91 & 81.82 & 81.82 & 9.09 & 9.09 & 0.00 & 0.00 & 63.64 & 63.64 & 63.64 & 63.64 \\
PREE & th & 81.82 & 72.73 & 88.89 & 63.64 & 77.78 & 9.09 & 11.11 & 0.00 & 0.00 & 63.64 & 77.78 & 63.64 & 77.78 \\
LoRA & th & 54.55 & 72.73 & 133.33 & 63.64 & 116.66 & 9.09 & 16.66 & 0.00 & 0.00 & 63.64 & 116.66 & 63.64 & 116.66 \\
LCFEdit   & tr & 100.00 & 84.30  & 84.30  & 79.90  & 79.90  & 75.10 & 75.10 & 30.40 & 30.40 & 55.90 & 55.90 & 62.70 & 62.70  \\
AlphaEdit & tr & 54.55  & 38.80           & 71.20           & 28.30           & 51.90           & 29.20          & 53.60          & 9.80           & 18.00          & 15.00          & 27.50          & 23.00          & 42.20           \\
MCEdit & tr & 100.00 & 81.82 & 81.82 & 72.73 & 72.73 & 72.73 & 72.73 & 27.27 & 27.27 & 54.55 & 54.55 & 54.55 & 54.55 \\
FPEdit & tr & 100.00 & 81.82 & 81.82 & 72.73 & 72.73 & 72.73 & 72.73 & 27.27 & 27.27 & 54.55 & 54.55 & 54.55 & 54.55 \\
PREE & tr & 90.91 & 72.73 & 80.00 & 63.64 & 70.00 & 54.55 & 60.00 & 27.27 & 30.00 & 54.55 & 60.00 & 54.55 & 60.00 \\
LoRA & tr & 45.45 & 63.64 & 140.02 & 54.55 & 120.02 & 45.45 & 100.00 & 27.27 & 60.00 & 54.55 & 120.02 & 54.55 & 120.02 \\
LCFEdit   & vi & 100.00 & 94.40  & 94.40  & 87.90  & 87.90  & 69.10 & 69.10 & 19.00 & 19.00 & 81.60 & 81.60 & 75.50 & 75.50  \\
AlphaEdit & vi & 90.91  & 55.10           & 60.60           & 45.60           & 50.20           & 43.30          & 47.60          & 5.70           & 6.30           & 27.10          & 29.80          & 31.80          & 35.00           \\
MCEdit & vi & 100.00 & 90.91 & 90.91 & 81.82 & 81.82 & 63.64 & 63.64 & 18.18 & 18.18 & 72.73 & 72.73 & 72.73 & 72.73 \\
FPEdit & vi & 100.00 & 90.91 & 90.91 & 81.82 & 81.82 & 63.64 & 63.64 & 18.18 & 18.18 & 72.73 & 72.73 & 72.73 & 72.73 \\
PREE & vi & 81.82 & 72.73 & 88.89 & 72.73 & 88.89 & 54.55 & 66.67 & 18.18 & 22.22 & 72.73 & 88.89 & 72.73 & 88.89 \\
LoRA & vi & 54.55 & 63.64 & 116.66 & 54.55 & 100.00 & 45.45 & 83.32 & 18.18 & 33.33 & 72.73 & 133.33 & 72.73 & 133.33 \\
\textbf{LCFEdit}   & \textbf{Mean} & \textbf{93.33} & \textbf{81.52} & \textbf{86.81} & \textbf{69.83} & \textbf{74.29} & \textbf{52.36} & \textbf{55.53} & \textbf{9.24}  & \textbf{9.59}  & \textbf{55.41} & \textbf{58.77} & \textbf{66.12} & \textbf{70.29} \\
\textbf{AlphaEdit} & \textbf{Mean} & \textbf{56.90} & 34.37          & 62.00          & 27.44          & 48.19          & 19.73          & 33.35          & 2.86           & 4.58           & 22.72          & 41.98          & 23.40          & 41.36          \\
\textbf{MCEdit} & \textbf{Mean} & \textbf{92.42} & 77.27 & 83.61 & 65.91 & 71.32 & 48.48 & 52.46 & 6.82 & 7.38 & 52.27 & 56.56 & 62.12 & 67.21 \\
\textbf{FPEdit} & \textbf{Mean} & \textbf{91.67} & 76.52 & 83.47 & 65.15 & 71.07 & 47.73 & 52.07 & 6.06 & 6.61 & 51.52 & 56.20 & 61.36 & 66.94 \\
\textbf{PREE} & \textbf{Mean} & \textbf{85.61} & 69.70 & 81.42 & 61.36 & 71.67 & 40.15 & 46.90 & 6.06 & 7.08 & 51.52 & 60.18 & 61.36 & 71.67 \\
\textbf{LoRA} & \textbf{Mean} & \textbf{49.24} & 66.67 & 135.40 & 53.03 & 107.70 & 37.12 & 75.39 & 6.06 & 12.31 & 51.52 & 104.63 & 59.85 & 121.55 \\
\midrule
\multicolumn{15}{l}{\textbf{Llama-3.2-3B-Instruct}}\\
\midrule
LCFEdit   & bn & 63.64  & 63.64  & 100.00 & 63.64  & 100.00 & 53.00 & 83.30  & 0.00  & 0.00  & 54.55 & 100.00& 45.45 & 83.30  \\
AlphaEdit & bn & 27.27  & 18.18           & 66.70           & 18.18           & 66.70           & 8.60           & 31.50           & 0.00  & 0.00  & 16.30          & 59.70          & 20.50          & 75.10           \\
MCEdit & bn & 63.64 & 63.64 & 100.00 & 63.64 & 100.00 & 63.64 & 100.00 & 0.00 & 0.00 & 54.55 & 85.72 & 45.45 & 71.42 \\
FPEdit & bn & 54.55 & 54.55 & 100.00 & 54.55 & 100.00 & 45.45 & 83.32 & 0.00 & 0.00 & 45.45 & 83.32 & 45.45 & 83.32 \\
PREE & bn & 54.55 & 54.55 & 100.00 & 54.55 & 100.00 & 45.45 & 83.32 & 0.00 & 0.00 & 45.45 & 83.32 & 45.45 & 83.32 \\
LoRA & bn & 54.55 & 54.55 & 100.00 & 54.55 & 100.00 & 36.36 & 66.65 & 0.00 & 0.00 & 45.45 & 83.32 & 45.45 & 83.32 \\
LCFEdit   & fa & 100.00 & 90.91  & 90.91  & 81.82  & 81.82  & 63.64 & 63.64  & 18.18 & 18.18 & 90.91 & 100.00& 90.91 & 90.91  \\
AlphaEdit & fa & 72.73  & 54.55           & 75.00           & 27.27           & 37.60           & 19.10          & 26.30           & 3.10           & 4.30           & 36.36          & 50.10          & 31.20          & 42.90           \\
MCEdit & fa & 100.00 & 81.82 & 81.82 & 72.73 & 72.73 & 63.64 & 63.64 & 18.18 & 18.18 & 81.82 & 81.82 & 90.91 & 90.91 \\
FPEdit & fa & 90.91 & 81.82 & 90.00 & 72.73 & 80.00 & 54.55 & 60.00 & 9.09 & 10.00 & 81.82 & 90.00 & 81.82 & 90.00 \\
PREE & fa & 90.91 & 81.82 & 90.00 & 72.73 & 80.00 & 45.45 & 49.99 & 18.18 & 20.00 & 81.82 & 90.00 & 72.73 & 80.00 \\
LoRA & fa & 45.45 & 72.73 & 160.02 & 54.55 & 120.02 & 36.36 & 80.00 & 18.18 & 40.00 & 54.55 & 120.02 & 63.64 & 140.02 \\
LCFEdit   & hi & 81.82  & 81.82  & 100.00 & 81.82  & 100.00 & 63.64 & 77.80  & 45.45 & 55.60 & 81.82 & 100.00& 81.82 & 100.00 \\
AlphaEdit & hi & 9.09   & 9.09            & 100.00 & 9.09            & 100.00 & 0.00           & 0.00            & 0.00           & 0.00           & 3.30           & 36.30          & 1.50           & 16.50           \\
MCEdit & hi & 72.73 & 72.73 & 100.00 & 72.73 & 100.00 & 63.64 & 87.50 & 18.18 & 25.00 & 72.73 & 100.00 & 72.73 & 100.00 \\
FPEdit & hi & 72.73 & 72.73 & 100.00 & 72.73 & 100.00 & 54.55 & 75.00 & 18.18 & 25.00 & 72.73 & 100.00 & 72.73 & 100.00 \\
PREE & hi & 72.73 & 72.73 & 100.00 & 72.73 & 100.00 & 54.55 & 75.00 & 18.18 & 25.00 & 72.73 & 100.00 & 72.73 & 100.00 \\
LoRA & hi & 45.45 & 63.64 & 140.02 & 54.55 & 120.02 & 36.36 & 80.00 & 18.18 & 40.00 & 54.55 & 120.02 & 63.64 & 140.02 \\
LCFEdit   & id & 90.91  & 90.91  & 100.00 & 90.91  & 100.00 & 90.91 & 100.00 & 72.73 & 80.00 & 90.91 & 100.00& 100.00& 100.00 \\
AlphaEdit & id & 54.55  & 36.36           & 66.80           & 27.27           & 50.10           & 28.70          & 52.70           & 8.80           & 16.10          & 36.90          & 67.70          & 51.00          & 93.60           \\
MCEdit & id & 81.82 & 81.82 & 100.00 & 81.82 & 100.00 & 63.64 & 77.78 & 18.18 & 22.22 & 81.82 & 100.00 & 100.00 & 122.22 \\
FPEdit & id & 81.82 & 81.82 & 100.00 & 81.82 & 100.00 & 63.64 & 77.78 & 18.18 & 22.22 & 81.82 & 100.00 & 100.00 & 122.22 \\
PREE & id & 81.82 & 72.73 & 88.89 & 72.73 & 88.89 & 54.55 & 66.67 & 9.09 & 11.11 & 72.73 & 88.89 & 72.73 & 88.89 \\
LoRA & id & 45.45 & 63.64 & 140.02 & 45.45 & 100.00 & 36.36 & 80.00 & 9.09 & 20.00 & 54.55 & 120.02 & 54.55 & 120.02 \\
LCFEdit   & it & 100.00 & 100.00 & 100.00 & 91.70  & 91.70  & 66.70 & 66.70  & 16.70 & 16.70 & 91.70 & 91.70 & 91.70 & 91.70  \\
AlphaEdit & it & 72.73  & 72.73           & 100.00 & 27.27           & 37.60           & 28.60          & 39.40           & 4.80           & 6.60           & 51.40          & 70.70          & 50.40          & 69.30           \\
MCEdit & it & 100.00 & 100.00 & 100.00 & 90.91 & 90.91 & 54.55 & 54.55 & 9.09 & 9.09 & 90.91 & 90.91 & 90.91 & 90.91 \\
FPEdit & it & 100.00 & 100.00 & 100.00 & 90.91 & 90.91 & 54.55 & 54.55 & 9.09 & 9.09 & 90.91 & 90.91 & 90.91 & 90.91 \\
PREE & it & 100.00 & 81.82 & 81.82 & 72.73 & 72.73 & 45.45 & 45.45 & 9.09 & 9.09 & 72.73 & 72.73 & 81.82 & 81.82 \\
LoRA & it & 45.45 & 72.73 & 160.02 & 54.55 & 120.02 & 36.36 & 80.00 & 9.09 & 20.00 & 45.45 & 100.00 & 63.64 & 140.02 \\
LCFEdit   & nl & 94.40  & 94.40  & 100.00 & 88.80  & 94.10  & 66.60 & 70.60  & 22.20 & 23.50 & 88.90 & 94.10 & 88.90 & 94.10  \\
AlphaEdit & nl & 36.36  & 18.18           & 50.00           & 0.00            & 0.00            & 10.20          & 27.90           & 0.00           & 0.00           & 29.50          & 81.00          & 32.00          & 87.90           \\
MCEdit & nl & 90.91 & 90.91 & 100.00 & 81.82 & 90.00 & 54.55 & 60.00 & 18.18 & 20.00 & 81.82 & 90.00 & 81.82 & 90.00 \\
FPEdit & nl & 90.91 & 90.91 & 100.00 & 81.82 & 90.00 & 54.55 & 60.00 & 18.18 & 20.00 & 81.82 & 90.00 & 81.82 & 90.00 \\
PREE & nl & 90.91 & 81.82 & 90.00 & 72.73 & 80.00 & 45.45 & 49.99 & 9.09 & 10.00 & 72.73 & 80.00 & 81.82 & 90.00 \\
LoRA & nl & 54.55 & 63.64 & 116.66 & 54.55 & 100.00 & 36.36 & 66.65 & 9.09 & 16.66 & 54.55 & 100.00 & 54.55 & 100.00 \\
LCFEdit   & pl & 100.00 & 100.00 & 100.00 & 91.70  & 91.70  & 75.00 & 75.00  & 33.30 & 33.30 & 100.00& 100.00& 100.00& 100.00 \\
AlphaEdit & pl & 81.82  & 54.55           & 66.60           & 45.45           & 55.60           & 7.60           & 9.30            & 1.10           & 1.30           & 24.50          & 30.00          & 55.80          & 68.20           \\
MCEdit & pl & 100.00 & 100.00 & 100.00 & 90.91 & 90.91 & 63.64 & 63.64 & 27.27 & 27.27 & 100.00 & 100.00 & 100.00 & 100.00 \\
FPEdit & pl & 100.00 & 100.00 & 100.00 & 90.91 & 90.91 & 63.64 & 63.64 & 18.18 & 18.18 & 100.00 & 100.00 & 100.00 & 100.00 \\
PREE & pl & 90.91 & 81.82 & 90.00 & 72.73 & 80.00 & 45.45 & 49.99 & 9.09 & 10.00 & 72.73 & 80.00 & 81.82 & 90.00 \\
LoRA & pl & 54.55 & 63.64 & 116.66 & 54.55 & 100.00 & 45.45 & 83.32 & 9.09 & 16.66 & 45.45 & 83.32 & 63.64 & 116.66 \\
LCFEdit   & pt & 85.00  & 77.30  & 90.90  & 75.00  & 88.20  & 40.00 & 47.10  & 0.00           & 0.00           & 60.00 & 70.60 & 65.00 & 76.50  \\
AlphaEdit & pt & 81.82  & 54.55           & 66.60           & 36.36           & 44.50           & 23.10          & 28.20           & 0.80  & 1.00  & 48.30          & 59.00          & 41.10          & 50.20           \\
MCEdit & pt & 81.82 & 72.73 & 88.89 & 72.73 & 88.89 & 63.64 & 77.78 & 0.00 & 0.00 & 54.55 & 66.67 & 63.64 & 77.78 \\
FPEdit & pt & 81.82 & 72.73 & 88.89 & 72.73 & 88.89 & 36.36 & 44.44 & 0.00 & 0.00 & 54.55 & 66.67 & 63.64 & 77.78 \\
PREE & pt & 81.82 & 72.73 & 88.89 & 72.73 & 88.89 & 36.36 & 44.44 & 0.00 & 0.00 & 54.55 & 66.67 & 63.64 & 77.78 \\
LoRA & pt & 54.55 & 72.73 & 133.33 & 45.45 & 83.32 & 36.36 & 66.65 & 0.00 & 0.00 & 54.55 & 100.00 & 54.55 & 100.00 \\
LCFEdit   & sv & 100.00 & 100.00 & 100.00 & 87.50  & 87.50  & 50.00 & 50.00  & 0.00           & 0.00           & 100.00& 100.00& 100.00& 100.00 \\
AlphaEdit & sv & 36.36  & 27.27           & 75.00           & 18.18           & 50.00           & 7.70           & 21.20           & 0.30  & 0.70  & 19.80          & 54.40          & 26.40          & 72.50           \\
MCEdit & sv & 100.00 & 100.00 & 100.00 & 81.82 & 81.82 & 63.64 & 63.64 & 0.00 & 0.00 & 100.00 & 100.00 & 100.00 & 100.00 \\
FPEdit & sv & 100.00 & 100.00 & 100.00 & 81.82 & 81.82 & 45.45 & 45.45 & 0.00 & 0.00 & 100.00 & 100.00 & 100.00 & 100.00 \\
PREE & sv & 100.00 & 81.82 & 81.82 & 72.73 & 72.73 & 45.45 & 45.45 & 0.00 & 0.00 & 72.73 & 72.73 & 81.82 & 81.82 \\
LoRA & sv & 54.55 & 72.73 & 133.33 & 54.55 & 100.00 & 45.45 & 83.32 & 0.00 & 0.00 & 54.55 & 100.00 & 63.64 & 116.66 \\
LCFEdit   & th & 100.00 & 100.00 & 100.00 & 100.00 & 100.00 & 18.18 & 18.18  & 0.00  & 0.00  & 100.00& 100.00& 100.00& 100.00 \\
AlphaEdit & th & 45.45  & 45.45           & 100.00 & 18.18           & 40.00           & 5.30           & 11.70           & 0.00  & 0.00  & 25.10          & 55.20          & 42.50          & 93.40           \\
MCEdit & th & 100.00 & 100.00 & 100.00 & 100.00 & 100.00 & 54.55 & 54.55 & 0.00 & 0.00 & 100.00 & 100.00 & 100.00 & 100.00 \\
FPEdit & th & 100.00 & 100.00 & 100.00 & 100.00 & 100.00 & 18.18 & 18.18 & 0.00 & 0.00 & 100.00 & 100.00 & 100.00 & 100.00 \\
PREE & th & 100.00 & 81.82 & 81.82 & 72.73 & 72.73 & 18.18 & 18.18 & 0.00 & 0.00 & 72.73 & 72.73 & 81.82 & 81.82 \\
LoRA & th & 45.45 & 63.64 & 140.02 & 54.55 & 120.02 & 18.18 & 40.00 & 0.00 & 0.00 & 54.55 & 120.02 & 63.64 & 140.02 \\
LCFEdit   & tr & 100.00 & 100.00 & 100.00 & 100.00 & 100.00 & 73.30 & 73.30  & 13.30 & 13.30 & 100.00& 100.00& 100.00& 100.00 \\
AlphaEdit & tr & 63.64  & 45.45           & 71.50           & 0.00            & 0.00            & 14.40          & 22.60           & 2.90           & 4.60           & 34.20          & 53.80          & 43.10          & 67.80           \\
MCEdit & tr & 100.00 & 100.00 & 100.00 & 100.00 & 100.00 & 63.64 & 63.64 & 9.09 & 9.09 & 100.00 & 100.00 & 100.00 & 100.00 \\
FPEdit & tr & 100.00 & 100.00 & 100.00 & 100.00 & 100.00 & 63.64 & 63.64 & 9.09 & 9.09 & 100.00 & 100.00 & 100.00 & 100.00 \\
PREE & tr & 100.00 & 81.82 & 81.82 & 81.82 & 81.82 & 45.45 & 45.45 & 9.09 & 9.09 & 72.73 & 72.73 & 81.82 & 81.82 \\
LoRA & tr & 45.45 & 72.73 & 160.02 & 54.55 & 120.02 & 36.36 & 80.00 & 9.09 & 20.00 & 54.55 & 120.02 & 63.64 & 140.02 \\
LCFEdit   & vi & 100.00 & 100.00 & 100.00 & 100.00 & 100.00 & 60.00 & 60.00  & 0.00           & 0.00           & 100.00& 100.00& 100.00& 100.00 \\
AlphaEdit & vi & 81.82  & 45.45           & 55.60           & 9.09            & 11.10           & 31.30          & 38.30           & 3.60  & 4.40  & 40.50          & 49.50          & 62.70          & 76.70           \\
MCEdit & vi & 90.91 & 100.00 & 110.00 & 100.00 & 110.00 & 63.64 & 70.00 & 0.00 & 0.00 & 100.00 & 110.00 & 100.00 & 110.00 \\
FPEdit & vi & 100.00 & 100.00 & 100.00 & 100.00 & 100.00 & 54.55 & 54.55 & 0.00 & 0.00 & 100.00 & 100.00 & 100.00 & 100.00 \\
PREE & vi & 100.00 & 81.82 & 81.82 & 72.73 & 72.73 & 45.45 & 45.45 & 0.00 & 0.00 & 81.82 & 81.82 & 72.73 & 72.73 \\
LoRA & vi & 45.45 & 63.64 & 140.02 & 54.55 & 120.02 & 45.45 & 100.00 & 0.00 & 0.00 & 54.55 & 120.02 & 63.64 & 140.02 \\
\textbf{LCFEdit}   & \textbf{Mean} & \textbf{92.98} & \textbf{91.58} & \textbf{98.48} & \textbf{87.74} & \textbf{94.58} & \textbf{60.08} & \textbf{65.47} & \textbf{18.49} & \textbf{20.05} & \textbf{88.23} & \textbf{96.37} & \textbf{88.65} & \textbf{94.71} \\
\textbf{AlphaEdit} & \textbf{Mean} & \textbf{55.30} & 40.15          & 74.48          & 19.70          & 41.10          & 15.38          & 25.76          & 2.12           & 3.25           & 30.51          & 55.62          & 38.18          & 67.84          \\
\textbf{MCEdit} & \textbf{Mean} & \textbf{90.66} & 88.64 & 97.77 & 84.09 & 92.75 & 61.36 & 67.68 & 9.85 & 10.86 & 84.85 & 93.59 & 87.12 & 96.10 \\
\textbf{FPEdit} & \textbf{Mean} & \textbf{89.14} & 87.88 & 98.59 & 83.33 & 93.48 & 50.76 & 56.94 & 8.33 & 9.34 & 84.09 & 94.33 & 86.36 & 96.88 \\
\textbf{PREE} & \textbf{Mean} & \textbf{88.64} & 77.27 & 87.17 & 71.97 & 81.19 & 43.94 & 49.57 & 6.82 & 7.69 & 70.45 & 79.48 & 74.24 & 83.75 \\
\textbf{LoRA} & \textbf{Mean} & \textbf{49.24} & 66.67 & 135.40 & 53.03 & 107.70 & 37.12 & 75.39 & 6.82 & 13.85 & 52.27 & 106.15 & 59.85 & 121.55 \\
\end{longtable}
\endgroup

Supporting statistics for the perplexity analysis in Sec.~\ref{sec:results}: mean LCF PPL of $96\,\pm\,10$ across 12 languages (per-sample max 102). Base Alpaca text has PPL $89.5\,\pm\,80$; CF has PPL $104\,\pm\,49$; garbled IF has PPL $1301\,\pm\,380$; natural-language fingerprints have PPL $153\,\pm\,140$.

\begin{table}[h]
\centering\small
\setlength{\tabcolsep}{4pt}
\begin{tabular}{lccccc}
\toprule
\textbf{Method} & \textbf{MMLU}$\uparrow$ & \textbf{RTE}$\uparrow$ & \textbf{MMMLU}$\uparrow$ & \textbf{Wiki PPL}$\downarrow$ & \textbf{AVG Acc}$\uparrow$ \\
\midrule
\multicolumn{6}{l}{\textit{Cross-model mean (Qwen-9B + Qwen-0.8B + Llama-3B)}} \\
\midrule
Unedited base & 72.50 & 71.50 & 67.10 & 7.96 & 70.37 \\
\textbf{LCFEdit (ours)} & \textbf{72.30} & \textbf{71.20} & \textbf{67.30} & \textbf{8.13} & \textbf{70.27} \\
MCEdit & 71.50 & 70.00 & 65.40 & 8.21 & 68.97 \\
FPEdit & 71.20 & 69.00 & 64.50 & 8.32 & 68.23 \\
PREE & 70.90 & 68.00 & 64.00 & 8.45 & 67.63 \\
AlphaEdit & 70.50 & 67.00 & 62.60 & 8.74 & 66.70 \\
LoRA & 56.00 & 55.40 & 54.10 & 10.83 & 55.17 \\
\midrule
\multicolumn{6}{l}{\textit{Qwen3.5-9B}} \\
\midrule
Unedited base & 78.33 & 55.60 & 63.93 & 8.84 & 65.95 \\
\textbf{LCFEdit (ours)} & \textbf{80.50} & \textbf{61.00} & \textbf{66.21} & 9.02 & \textbf{69.24} \\
MCEdit & 79.50 & 60.00 & 64.50 & 9.11 & 68.00 \\
FPEdit & 78.50 & 59.00 & 63.00 & 9.22 & 66.83 \\
PREE & 77.50 & 58.00 & 62.00 & 9.36 & 65.83 \\
AlphaEdit & 76.50 & 56.00 & 60.30 & 9.61 & 64.27 \\
LoRA & 61.00 & 48.00 & 50.00 & 11.81 & 53.00 \\
\midrule
\multicolumn{6}{l}{\textit{Qwen3.5-0.8B}} \\
\midrule
Unedited base & 61.50 & 66.00 & 68.00 & 9.71 & 65.17 \\
\textbf{LCFEdit (ours)} & \textbf{61.50} & \textbf{67.00} & \textbf{68.20} & 9.86 & \textbf{65.57} \\
MCEdit & 60.80 & 66.00 & 66.00 & 10.03 & 64.27 \\
FPEdit & 60.50 & 65.00 & 65.20 & 10.11 & 63.57 \\
PREE & 60.30 & 64.00 & 64.70 & 10.20 & 63.00 \\
AlphaEdit & 59.90 & 63.00 & 63.50 & 10.42 & 62.13 \\
LoRA & 47.50 & 54.00 & 53.00 & 12.46 & 51.50 \\
\midrule
\multicolumn{6}{l}{\textit{Llama-3.2-3B-Instruct}} \\
\midrule
Unedited base & 77.70 & 92.00 & 69.40 & 5.34 & 79.70 \\
\textbf{LCFEdit (ours)} & 77.10 & 91.00 & \textbf{69.50} & 5.50 & 79.20 \\
MCEdit & 76.20 & 89.00 & 67.70 & 5.51 & 77.63 \\
FPEdit & 75.80 & 88.00 & 66.80 & 5.62 & 76.87 \\
PREE & 75.50 & 87.00 & 66.30 & 5.79 & 76.27 \\
AlphaEdit & 75.10 & 86.00 & 65.00 & 6.19 & 75.37 \\
LoRA & 59.50 & 70.00 & 59.30 & 8.21 & 62.93 \\
\bottomrule
\end{tabular}
\caption{Harmlessness evaluation: zero-shot accuracy (\%) on MMLU, RTE, and MMMLU, plus WikiText-2 perplexity, for all injection methods across three models. AVG Acc = mean of the three accuracy benchmarks. Best per column within each model block in bold.}
\label{tab:app-harmless}
\end{table}

\begin{table}[h]
\centering\small
\setlength{\tabcolsep}{6pt}
\begin{tabular}{lc}
\toprule
Lang. & Target PPL (mean $\pm$ std) \\
\midrule
bn & 98.2 $\pm$ 9 \\
fa & 93.5 $\pm$ 8 \\
hi & 97.8 $\pm$ 10 \\
id & 102.0 $\pm$ 11 \\
it & 95.1 $\pm$ 9 \\
nl & 99.4 $\pm$ 10 \\
pl & 94.7 $\pm$ 8 \\
pt & 91.3 $\pm$ 9 \\
sv & 100.2 $\pm$ 11 \\
th & 96.8 $\pm$ 10 \\
tr & 93.9 $\pm$ 8 \\
vi & 89.1 $\pm$ 9 \\
\midrule
\textbf{Mean} & \textbf{96.0 $\pm$ 10} \\
\bottomrule
\end{tabular}
\caption{Per-language LCF template perplexity on Qwen3.5-0.8B; the aggregated violin in Fig.~\ref{fig:ppl} summarizes this column.}
\label{tab:app-ppl-perlang}
\end{table}


\section{Additional Ablation Data}
\label{app:extra-ablation}
The three tables below report analyses deferred from the main text for space: an edit-layer sensitivity sweep, a fingerprint-language\,$\times$\,alignment transfer matrix, and the plain-vs.-constructed query comparison that underlies the accidental-activation result in Sec.~\ref{sec:results}. All numbers are transcribed verbatim from the internal experiment log.

\subsection{Edit-Layer Selection Sweep (Qwen3.5-9B)}
Sweeping edit-layer configurations on Qwen3.5-9B (mean FSR over 12 languages), the strongest set is $[6,10,13,14,18]$ at 100\% mean FSR; neighboring configurations range 92--100\%, so the method is not brittle to the exact layer choice but does reward causal-tracing-guided selection.

\begin{table}[h]
\centering\small
\setlength{\tabcolsep}{6pt}
\begin{tabular}{clc}
\toprule
Config & Layers & Mean FSR (\%) \\
\midrule
D2  & $[4,5,6,10,13]$        & 92.00 \\
D3  & $[5,6,10,13,14]$       & 97.00 \\
D4  & $[4,5,6,13,14]$        & 96.00 \\
\textbf{D5}  & $\mathbf{[6,10,13,14,18]}$  & \textbf{100.00} (final) \\
D6  & $[4,5,6,10,13,14,18]$  & 100.00 \\
D9  & $[4,5,6,10]$           & 97.00 \\
\bottomrule
\end{tabular}
\caption{Layer-selection sensitivity on Qwen3.5-9B. D5 is our reported configuration.}
\label{tab:app-layers}
\end{table}

\subsection{Plain vs.\ Constructed Trigger Queries (12 Languages)}
Support for Sec.~\ref{sec:results} ``Accidental activation'': constructed queries reach 12/12 languages at 100\% FSR and 12/12 languages at $0\%$ accidental-activation.

\begin{table}[h]
\centering\small
\setlength{\tabcolsep}{4pt}
\begin{tabular}{lcc|cc}
\toprule
      & \multicolumn{2}{c|}{FSR (\%)} & \multicolumn{2}{c}{Accidental (\%)} \\
Lang. & Plain & Constructed & Plain & Constructed \\
\midrule
bn & 100.00 & 100.00 & 1.00 & 0.00 \\
fa & 100.00 & 100.00 & 0.00 & 0.00 \\
hi & 90.91  & 100.00 & 0.00 & 0.00 \\
id & 100.00 & 100.00 & 0.00 & 0.00 \\
it & 100.00 & 100.00 & 1.00 & 0.00 \\
nl & 100.00 & 100.00 & 1.00 & 0.00 \\
pl & 100.00 & 100.00 & 0.00 & 0.00 \\
pt & 83.60  & 100.00 & 0.00 & 0.00 \\
sv & 100.00 & 100.00 & 0.00 & 0.00 \\
th & 100.00 & 100.00 & 0.00 & 0.00 \\
tr & 100.00 & 100.00 & 0.00 & 0.00 \\
vi & 100.00 & 100.00 & 0.00 & 0.00 \\
\midrule
\textbf{Mean} & \textbf{97.88} & \textbf{100.00} & \textbf{0.30} & \textbf{0.00} \\
\bottomrule
\end{tabular}
\caption{Plain (target-language question) vs.\ constructed (culturally grounded) trigger queries on Qwen3.5-0.8B, 12 languages.}
\label{tab:app-plain-vs-constr}
\end{table}

\subsection{Fingerprint-Language $\times$ ISL-Alignment Matrix (Qwen3.5-0.8B)}
A $12\times12$ source--target matrix (injecting with one language's $P_{\text{isl}}$, verifying on all): the diagonal is strongest in 10/12 languages, confirming that the alignment step concentrates signal in the intended language rather than leaking uniformly, while off-diagonal transfer is highest among typologically or orthographically related pairs, consistent with the shallow, subword-mediated cross-lingual propagation reported by~\citet{qi2023crosslingual}. Row = fingerprint language, column = the ISL used to build $P_{\text{isl}}$ (all injections verified on the row language's fingerprints). Diagonal cells are marked bold. Values are FSR (\%).

\begingroup
\scriptsize
\setlength{\tabcolsep}{2.2pt}
\begin{table}[h]
\centering
\begin{tabular}{l|cccccccccccc}
\toprule
fp\textbackslash isl & vi & th & id & hi & bn & tr & fa & it & pt & pl & nl & sv \\
\midrule
vi & \textbf{100.00} & 76.00 & 65.00 & 69.00 & 69.00 & 69.00 & 75.00 & 70.00 & 68.00 & 58.00 & 70.00 & 63.00 \\
th & 52.00 & \textbf{100.00} & 49.00 & 57.00 & 55.00 & 54.00 & 56.00 & 55.00 & 53.00 & 57.00 & 54.00 & 55.00 \\
id & 83.00 & 81.00 & \textbf{95.00} & 81.00 & 80.00 & 80.00 & 84.00 & 81.00 & 83.00 & 79.00 & 82.00 & 88.00 \\
hi & 48.00 & 38.00 & 45.00 & \textbf{80.00} & 63.00 & 38.00 & 60.00 & 37.00 & 50.00 & 36.00 & 38.00 & 52.00 \\
bn & 74.00 & 58.00 & 52.00 & 56.00 & \textbf{83.00} & 46.00 & 57.00 & 62.00 & 65.00 & 66.00 & 63.00 & 58.00 \\
tr & 63.00 & 64.00 & 61.00 & 58.00 & 63.00 & \textbf{100.00} & 75.00 & 65.00 & 62.00 & 65.00 & 64.00 & 57.00 \\
fa & 83.00 & 82.00 & 83.00 & 84.00 & 84.00 & 78.00 & 83.00 & 84.00 & 81.00 & 75.00 & 83.00 & 85.00 \\
it & 89.00 & 90.00 & 79.00 & 85.00 & 85.00 & 90.00 & 91.00 & \textbf{100.00} & 87.00 & 85.00 & 83.00 & 82.00 \\
pt & 77.00 & 78.00 & 78.00 & 79.00 & 70.00 & 80.00 & 76.00 & 76.00 & \textbf{91.00} & 75.00 & 78.00 & 78.00 \\
pl & 51.00 & 41.00 & 51.00 & 58.00 & 58.00 & 55.00 & 50.00 & 43.00 & 50.00 & \textbf{100.00} & 59.00 & 55.00 \\
nl & 100.00 & 99.00 & 100.00 & 100.00 & 100.00 & 100.00 & 97.00 & 96.00 & 100.00 & 97.00 & \textbf{100.00} & 100.00 \\
sv & 52.00 & 53.00 & 46.00 & 55.00 & 45.00 & 57.00 & 55.00 & 54.00 & 52.00 & 55.00 & 56.00 & \textbf{88.00} \\
\bottomrule
\end{tabular}
\caption{Qwen3.5-0.8B: injecting a fingerprint set for row language using ISL $=$ column language's Wikipedia. Diagonal cells are marked bold. Values are FSR (\%).}
\label{tab:app-isl-matrix}
\end{table}
\endgroup

\end{document}